%% file: main.tex
\RequirePackage{rotating}
\documentclass[journal]{IEEEtran}
\usepackage{booktabs}
\usepackage{textcomp, gensymb}
\usepackage{makecell}
\usepackage{tabularx}
\usepackage{multirow}
\usepackage{lipsum}
\usepackage{enumitem}
\usepackage{calc}
\usepackage{bm}
\usepackage{graphicx}
\usepackage{caption}
\usepackage{float}
\usepackage[export]{adjustbox}
\usepackage[caption = false]{subfig}
\usepackage{calrsfs}
\usepackage{float}
\usepackage{mathtools}
\usepackage{xcolor}
\usepackage{colortbl}
\usepackage[symbol]{footmisc}

\usepackage{amsmath,amssymb}
\usepackage{url}
\usepackage{bbm}
\usepackage{array}
\usepackage{longtable}

\makeatletter
\newcommand*{\rom}[1]{\expandafter\@slowromancap\romannumeral #1@}
\makeatother
\DeclareMathOperator*{\argmax}{argmax}

\DeclareMathAlphabet\mathbfcal{OMS}{cmsy}{b}{n}
\AtBeginDocument{%
 \providecommand\BibTeX{{%
  \normalfont B\kern-0.5em{\scshape i\kern-0.25em b}\kern-0.8em\TeX}}}

\begin{document}

\title{Design Automation for Fast, Lightweight, and Effective Deep Learning Models: A Survey}

\author{
Dalin Zhang,~\IEEEmembership{Member,~IEEE,}
Kaixuan Chen,~\IEEEmembership{Member,~IEEE,}
Yan Zhao,~\IEEEmembership{Member,~IEEE,}   
Bin Yang,~\IEEEmembership{Senior Member,~IEEE,}
Lina Yao,~\IEEEmembership{Senior Member,~IEEE,}
and Christian S. Jensen,~\IEEEmembership{Fellow,~IEEE}

\thanks{Dalin Zhang, Kaixuan Chen, Yan Zhao, Bin Yang, and Christian S. Jensen is with the Department of Computer Science, Aalborg University, Aalborg Øst 9220, Denmark (e-mail: dalinz@cs.aau.dk, kchen@cs.aau.dk, yanz@cs.aau.dk, byang@cs.aau.dk, and csj@cs.aau.dk).}
\thanks{Lina Yao is with the School of Computer Science and Engineering, University of New South Wales, UNSW Sydney 2052, Australia (e-mail: lina.yao@unsw.edu.au).}
\thanks{Manuscript received March 10, 2022}
}

\markboth{IEEE Communications Surveys \& Tutorials,~Vol.~14, No.~8, August~2015}%
{Zhang \MakeLowercase{\textit{et al.}}: Design Automation for Fast, Lightweight, and
Effective Deep Learning Models: A Survey}
 
\maketitle

\begin{abstract}
Deep learning technologies have demonstrated remarkable effectiveness in a wide range of tasks, and deep learning holds the potential to advance a multitude of applications, including in edge computing, where deep models are deployed on edge devices to enable instant data processing and response. A key challenge is that while the application of deep models often incurs substantial memory and computational costs, edge devices typically offer only very limited storage and computational capabilities that may vary substantially across devices. These characteristics make it difficult to build deep learning solutions that unleash the potential of edge devices while complying with their constraints.
A promising approach to addressing this challenge is to automate the design of effective deep learning models that are lightweight, require only a little storage, and incur only low computational overheads. This survey offers comprehensive coverage of studies of design automation techniques for deep learning models targeting edge computing. It offers an overview and comparison of key metrics that are used commonly to quantify the proficiency of models in terms of effectiveness, lightness, and computational costs. The survey then proceeds to cover three categories of the state-of-the-art of deep model design automation techniques: automated neural architecture search, automated model compression, and joint automated design and compression. Finally, the survey covers open issues and directions for future research.

\end{abstract}

\begin{IEEEkeywords}
deep learning, neural architecture search, lightweight model, model compression
\end{IEEEkeywords}

\input{content/1.introduction}
\input{content/2.evaluation_metrics}
\input{content/3.automated_search}
\input{content/4.automated_compression}
\input{content/5.joint}
\input{content/6.future}
\input{content/7.conclusion}

\bibliographystyle{IEEEtran}
\bibliography{ref-short}

\end{document}

%% file: content/1.introduction.tex
\section{Introduction}
\subsection{Background}
Deep learning has achieved state-of-the-art performance at a multitude of tasks and has affected people's lives in myriad areas, including recommendation systems \cite{DBLP:journals/csur/ZhangYST19}, natural language understanding \cite{jacob2019bert}, and biomedical engineering \cite{zhang2019learning}. Deep learning frees researchers from manually designing purposeful feature representations of objects by introducing multi-layer neural architectures capable of automatic feature extraction. This enables researchers to work at a higher level of abstraction, focusing on \textit{architecture engineering} rather than on feature engineering and model building. The neural architectures of deep models tend to be increasingly intricate and complex, thus requiring substantial hardware resources for deployment. This may not be a problem when a powerful server is available, but important settings occur where this is not the case:: 1) computing on mobile hardware, e.g., smartphones and tablet PCs; 2) computing on industrial hardware optimized for low deployment cost and low power consumption. 
Furthermore, although cloud computing may be available, it is often fundamentally unattractive to transfer data from edge devices to the cloud. On the one hand, such transfer may incur privacy, ownership, and consequent regulatory concerns, including for human-related data like audio and video data and utility consumption data \cite{chen2012data}; on the other hand, data transfer incurs substantial latency due to low bandwidth ``last mile'' connectivity. Last but not least, deployment of complex models is expensive, due to the cost of specialized hardware and energy consumption, and it is also bad for the environment due to the carbon footprint of producing the required electricity \cite{strubell2019energy,schwartz2020green}.

Fortunately, research has demonstrated that deep learning models generally have large numbers of redundant parameters and computations that contribute to their performance \cite{DBLP:conf/nips/DenilSDRF13,LeCun1989nips}, so there is considerable room for reducing redundancies without compromising model accuracy. 
Hence, it is highly desirable and completely possible to design simplified deep learning models with reduced computational complexity, thus achieving lighter weight and more efficient deep models\footnote{By efficient deep learning models we mean deep learning models with low memory usage or low inference cost or latency.}.

\subsection{Design of Efficient Deep Learning Models}
Research on building efficient deep learning models can be categorized into two categories: \textit{efficient network architecture design} and \textit{model compression}. In \textbf{efficient network architecture design}, the aim is to create compact neural modules and to connect these according to a carefully designed topology. The objective is to achieve efficient deep learning models with acceptable accuracy but small structures (low memory requirement) and low computational complexity (high speed). MobileNet \cite{howard2017mobilenets} proposes an efficient network structure using a depthwise separable convolution module as the basic building block, and superior size, speed, and accuracy characteristics over a variety of computer vision tasks are documented. A second version, called MobileNetV2, incorporates a new basic building block, bottleneck depth-separable convolution with residuals/inverted residual \cite{mobilenetv2}. As a result, MobileNetV2 generally needs 30\% fewer parameters, requires two times fewer operations and is about 30--40\% faster on a Google Pixel phone while achieving higher accuracy than its predecessor. In ShuffleNetV1 \cite{zhang2018shufflenet}, bottleneck-like structures with pointwise group convolutions and "channel shuffle" operations are the basic building blocks that are used to achieve efficient neural structures. Like in the MobileNet case, ShuffleNet also has a second generation, called ShuffleNetV2 \cite{Ma_2018_ECCV}. Here a new ``channel split'' operation is introduced in order to further improve speed and accuracy. In addition, four practical guidelines for efficient network architecture design are provided. 

In contrast to directly designing efficient architecture from a pool of basic building blocks, \textbf{model compression} aims at modifying a given neural model to reduce its memory and computational cost. \textit{Pruning} is one such powerful technique that tries to remove unimportant components from a model \cite{he2017channel,luo2017thinet}. It is flexible in that it is possible to remove layers, neurons, connections, or channels. While pruning shrinks a model by removing redundant parts, \textit{quantization} aims to reduce the number of bits required to represent model parameters \cite{courbariaux2014training}. Most processors use 32 bits or more to store the parameters of a deep model. However, research estimates that the human brain stores information in a discrete format that uses 4--7 bits \cite{mcculloch1943logical,bartol2015hippocampal,linden2018think}. Indeed, many efforts have been devoted to investigating using fewer bits to store model parameters to reduce memory and computational cost \cite{lin2016fixed,gupta2015deep}. Other techniques like knowledge distillation \cite{hinton2015distilling} and tensor decomposition \cite{ye2018learning} are also popular and effective at compressing deep models.

\begin{figure*}
    \centering
    \includegraphics[width=1.7\columnwidth]{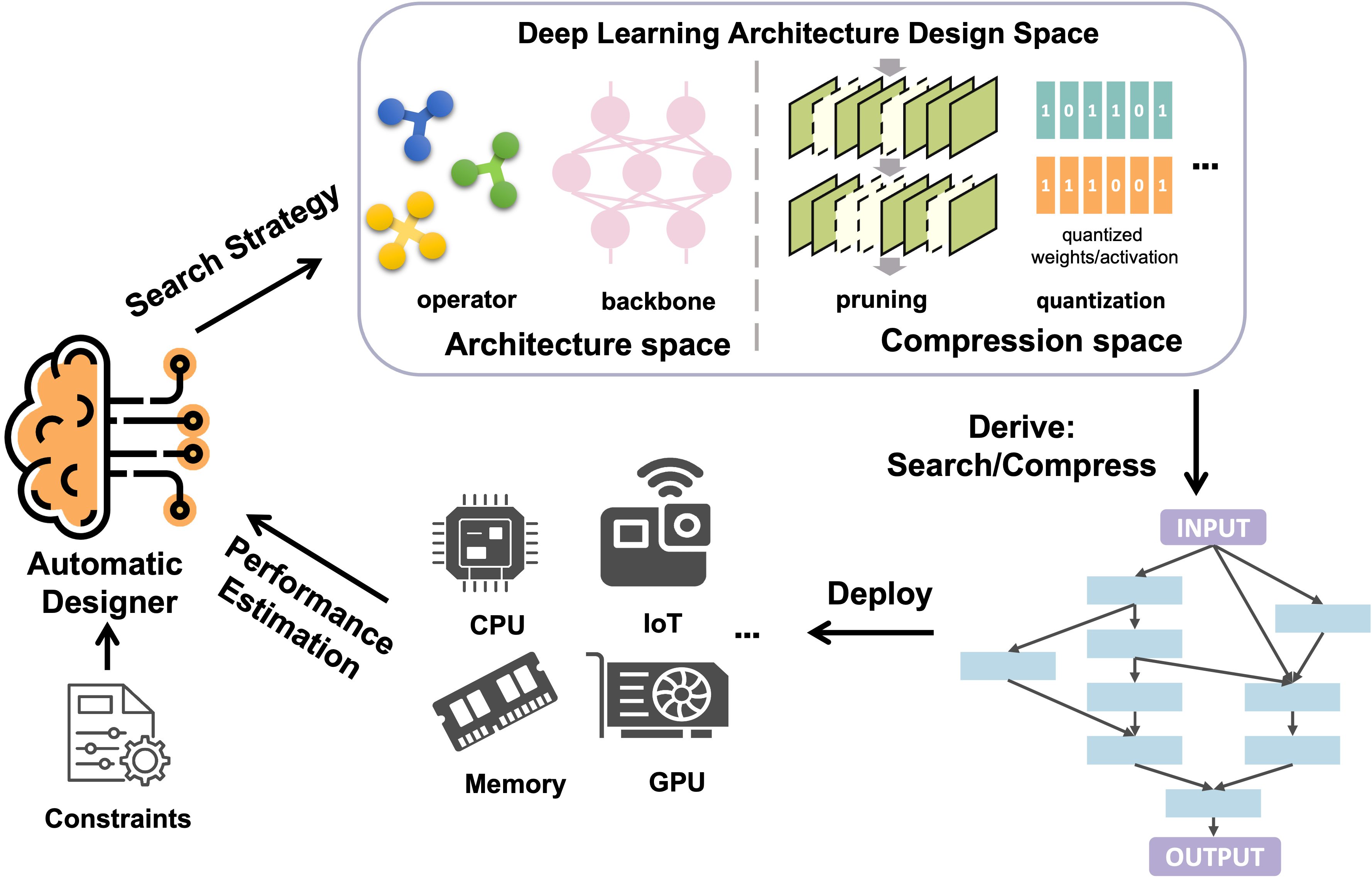}
    \caption{Design Automation for Efficient Deep Learning Models. The automatic designer takes the hardware constraints into its search strategy to explore a predefined deep learning architecture design space which could be either an architecture space or a compression space; with the explored design choices, a new deep learning model from scratch or a compressed deep learning model is derived and deployed on the target device; the model's performance including both accuracy and hardware consumption is finally estimated and fed into the automatic designer. }
    \label{fig:flow}
\end{figure*}

\subsection{Design Automation for Efficient Deep Learning Models}
Although remarkable progress has been achieved in building efficient deep learning models, initial proposals were heuristic rule-based and hand-tuned with inevitable limitations. First, building an efficient deep learning model still requires advanced prior knowledge and experience, making it difficult for beginners and even deep learning experts without domain knowledge to develop specialized models that meet given requirements. Second, as it is impossible to apply the same uniform model across diverse mobile platforms and tasks, enabling specializations that address such diversity is essential. Yet, it remains excessively time-consuming and inconvenient. Third, hand-crafted rules offer limited capabilities at utilizing hardware potentials fully, while satisfying the size and latency requirements. Different deep learning models can satisfy the same hardware constraints, and it is impractical to manually exhaust all possibilities. 

Observations such as the above have prompted a multitude of studies of design automation techniques for efficient deep learning models. Fig. \ref{fig:flow} shows a general automated design process. The design automation algorithm (i.e., \textit{Automatic Designer}) applies a search strategy to find network architectures and compression in the predefined \textit{Deep Learning Architecture Design Space}; next, a specific deep learning model is \textit{Derived} through executing the identified operations and is then \textit{Deployed} on target devices (e.g., CPU, GPU, or IoT devices); lastly, model performance metrics such as \textit{accuracy}, \textit{latency}, and \textit{memory use} are estimated and provided to the Automatic Designer. In the next iteration, the Automatic Designer considers both the feedback and specialized constraints and takes a new design action to find a better model in the design space. This process is repeated until a satisfactory model is achieved.

Neural architecture search (NAS) aims to automate the design of neural networks that achieve the best possible accuracy. Next, more targeted studies that aim to automate the design of efficient neural networks build on generic NAS and involve the design of search spaces and the modification of the optimization objective from sole accuracy to both accuracy and efficiency. 
Cai et al. \cite{proxyless} incorporate model latency into the optimization goal of their binarized design automation framework (i.e., ProxylessNAS) by means of a differentiable loss. Due to targeting optimized inference latency directly, ProxylessNAS can achieve efficient neural architectures 1.83 times faster than MobileNetV2 with the same level of top-1 accuracy. Notably, the search space of ProxylessNAS is based on the inverted residual blocks of different convolution sizes, as proposed by MobleNetV2. This implies that design automation is not only able to reduce human labor but even enables architectures that surpass handcrafted architectures. 

In addition to studies that target the direct design of efficient neural networks, other studies target the automated compression of deep neural networks \cite{xiao2019autoprune,wang2019haq, He_2018_ECCV}. For example, Xiao et al. \cite{xiao2019autoprune} design an automatic pruning approach that is based on learnable pruning indicators instead of pruning rules designed individually for specific architectures and datasets. This approach achieves superior compression performance on different widely-used neural models (e.g., AlexNet, ResNet, and MobileNet). Considering the progress in, and promise of, design automation for efficient deep learning models, it is a comprehensive and systematic survey is called for and holds the potential to accelerate future research. 

\subsection{Key Contributions} 
To the best of our knowledge, this is the first survey of state-of-the-art design automation methods that target fast, lightweight, and effective deep learning models. The key contributions of the survey are summarized as follows:
\begin{itemize}
    \item We provide a comprehensive review of design automation techniques targeting fast, lightweight, and effective deep learning models. In doing so, more than 150 papers are covered, analyzed, and compared.
    \item We propose a new taxonomy of deep design automation methods from the perspectives of how to design, i.e., by search (Section \ref{section:search}), by compression (Section \ref{section:compression}), or by joint search and compression (Section \ref{section:joint}); and by what to design, i.e., the search space, the search strategy, and the performance estimation strategy; and by what to compress, e.g., tensors, knowledge, and representation. The detailed categories provide convenience to the readers in obtaining an overview of the literature and identifying a direction of interest.
    \item We summarize and compare the evaluation metrics that are used for both the obtained models and the design approaches. By means of the comparison, we emphasize the role of each metric and explicate the associated pros and cons.
    \item We discuss open issues and identify future directions on automated design and compression.  
\end{itemize}

The remainder of the survey is organized as follows: Section~\ref{sec:evaluation} summarizes the evaluation metrics of efficiency (i.e., speed and lightness) and effectiveness of deep learning models. Section~\ref{section:search} covers studies of searching for efficient deep models.
Section~\ref{section:compression} then covers research on automated compression, and
Section~\ref{section:joint} considers studies of joint automated search and compression.
Section~\ref{sec:future} presents research directions, and Section~\ref{sec:conclusion} concludes the survey.

%% file: content/2.evaluation_metrics.tex
\section{Evaluation Metrics}\label{sec:evaluation}
Different evaluation metrics or objectives may lead to different or even opposite conclusions. Thus, it is indispensable to introduce and distinguish the relevant evaluation metrics before diving into the details of efficient models. In this section, we will introduce the evaluation metrics for measuring the efficiency and effectiveness of the \textit{obtained efficient models}. These metrics are also critical to evaluating the effectiveness of a design automation approach. Furthermore, we introduce metrics for evaluating the cost of a \textit{design automation approach} as well. We briefly summarize the characteristics including advantages and limitations of the commonly used evaluation metrics in Table \ref{tab:metrics}.

\begin{sidewaystable*}[ph]
\centering
\caption{Evaluation Metrics for Measuring the Efficiency and Effectiveness of Both the Obtained Efficient Model and the Design Automation Approach}
\begin{tabular}{p{0.16\linewidth}|c|p{0.3\linewidth}|p{0.3\linewidth}} 
\hline
\multicolumn{1}{l|}{\multirow{2}{*}{}}                  & \multirow{2}{*}{Metrics} & \multicolumn{2}{c}{Characteristics} \\ 
\cline{3-4}
\multicolumn{1}{l|}{}                                   &   & \multicolumn{1}{c|}{Advantages}                                                                                   & \multicolumn{1}{c}{Limitations} \\ 
\hline
\multirow{12}{\linewidth}{Device-agnostic Efficiency Evaluation Metrics} & \multirow{7}{*}{FLOPs}    & 
\vspace{.7cm}\begin{itemize}[leftmargin=0.5cm]
    \item each to obtain
    \item coarse approximation of latency
\end{itemize}
&
\begin{itemize}[leftmargin=0.5cm]
    \item inconsistent definitions
    \item omitting the degree of computational parallelism
    \item omitting the memory access cost
    \item omitting the implementation library
\end{itemize} \\

\cline{2-4}   & \multirow{5}{*}{Number of Parameters}     & 
              \begin{itemize}[leftmargin=0.5cm]
                   \item easy to obtain~
                   \item precise when measuring storage requirement
              \end{itemize} &    
              \begin{itemize}[leftmargin=0.5cm]
                  \item omitting the computation caching for measuring memory
                  \item unable to reflect peak memory usage
              \end{itemize}\\ 
\hline
\multirow{10}{\linewidth}{Device-aware Efficiency Evaluation Metrics}    & \multirow{5}{*}{Latency}                  & 
\multicolumn{2}{p{.6\linewidth}@{}}{
\begin{itemize}[leftmargin=0.5cm] 
        \item the real criterion that we care about
        \item obtain through implementation on real hardware or prediction models
        \item depend on different hardware platforms and implementation libraries
\end{itemize}}

\\

\cline{2-4}
                                                        & \multirow{5}{*}{Peak Memory Usage}            &
\multicolumn{2}{p{.6\linewidth}@{}}{
\begin{itemize}[leftmargin=0.5cm] 
        \item the real criterion that we care about
        \item obtain through implementation on real hardware or prediction models
        \item reflect the peak usage that really matters
\end{itemize}}
\\ 
\hline
\multirow{4}{\linewidth}{Effectiveness Evaluation Metrics}      & \multirow{4}{*}{Accuracy/mAP/mIOU etc.}   &
\multicolumn{2}{p{.6\linewidth}@{}}{
\begin{itemize}[leftmargin=0.5cm] 
        \item diverse and dependent on the targeting tasks
        \item identical to those for evaluating normal deep learning models
\end{itemize}}
\\ 
\hline
\multirow{8}{\linewidth}{Design Automation Cost Metrics}         & \multirow{4}{*}{GPU Hours}   &
\multicolumn{2}{p{.6\linewidth}@{}}{
\begin{itemize}[leftmargin=0.5cm] 
        \item used to evaluate the speed of a design automation method
        \item depend on the number of GPUs used
\end{itemize}}
\\ 
\cline{2-4}
 & \multirow{4}{*}{GPU Memory}               &
\multicolumn{2}{p{.6\linewidth}@{}}{
\begin{itemize}[leftmargin=0.5cm] 
        \item used to evaluate the memory requirement of a design automation
        \item grows linearly w.r.t. the size of the candidate set
\end{itemize}} 
\\
\hline
\end{tabular}
\label{tab:metrics}
\end{sidewaystable*}

\subsection{Device-agnostic Evaluation Metrics}
The device-agnostic metrics can be calculated directly from the model architecture without real-world implementation on hardware. These metrics do not essentially reflect a model's real performance that we care about \cite{Ma_2018_ECCV,langerman2020beyond,yao2018fastdeepiot}.

\subsubsection{FLoating-point OPerations (FLOPs)}
FLOPs is generally defined as \textit{the number of floating-point multiplication-add operations} in a model for approximating the latency/speed or computation complexity \cite{zhang2018shufflenet,Ma_2018_ECCV,yang2021condensenet}. There are also other commonly used analogous metrics, such as the number of multiply-add operations (MAdds) and the number of multiply-accumulate operations (MACs) \cite{camus2019survey,langerman2020beyond,chu2021discovering}. However, one contention remains regarding the definition: whether multiplication-add should be considered as one or two operations. Some researchers argue that in many recent deep learning models, convolutions are bias-free and it makes sense to count multiplication and add as separate FLOPs \cite{bianco2018benchmark}. Moreover, some non-multiplication or non-add operations require FLOPs in some implementations as well, such as an activation layer \cite{zhengbo2020research,langerman2020beyond}. Whether such operations should be counted into total FLOPs is also a dispute.

In addition to the inconsistency of the definition, there exist three main issues that induce the discrepancy between FLOPs and real latency. First, counting only FLOPs ignores some decisive factors that affect latency remarkably. One such factor is \textit{parallelism}. With the same FLOPs, a model with a high degree of parallelism may be much faster than another model with a low degree of parallelism  \cite{dryden2019improving,niu2020achieving}. Another important factor arises from the \textit{memory access cost}. Since the on-device RAM is usually limited, data cannot be entirely loaded at one time and thus reading data from external memory is required. It has an increasing impact on the latency as the computation unit is getting stronger recently, thus becoming the bottleneck for latency. This intrinsic factor should not be simply neglected.
Third, the implementation library of a deep learning model significantly influences its latency as well. For example, NVIDIA's cuDNN library provides different implementations of a convolution operation \cite{chetlur2014cudnn} that clearly require different amounts of FLOPs although the network architecture is identical. Some works also found that a smaller number of FLOPs could be even slower due to the library abstractions \cite{Ma_2018_ECCV,he2017channel}.

\subsubsection{Number of Parameters}
It is usually required to fit all parameters of a neural network within on-chip memory to execute the model fast \cite{liberis2019neural}. Thus, the number of parameters mainly constrains the memory requirement of a deep learning model. Although some mobile devices like smartphones have abundant memory, there are still a few mobile devices that have particularly scarce memory, such as microcontrollers that typically have 10’s-100’s of KB \cite{suda2019machine}. These memory-scarce devices are in demand in many fields due to their low prices. Thus, it is desirable to design a small-sized model that can fit into ``tiny'' memory and the number of parameters is a common device-agnostic metric for evaluating such a requirement. Nevertheless, the number of parameters only accounts for a portion of memory usage; input data, computation caching (i.e., intermediate tensors produced at runtime), and network structure information take over a relatively larger portion of memory \cite{siu2018memory,xu2018saving}. In some extreme scenarios, only computation caching and active parameters (i.e., the parameters used for current computation) occupy memory. Some researchers propose to optimize the in-memory computation caching to reduce memory consumption but their performance largely depends on the network structure \cite{unlu2020efficient}. Thus, a model with a larger number of parameters is not certainly more memory-hungry than a model with a smaller number of parameters. On the other hand, model parameters account for a major component of storage/external memory (e.g., FLASH memory) usage. 

\subsection{Device-aware Efficiency Evaluation Metrics}
Unlike device-agnostic metrics, device-aware metrics can reflect the computational cost that we really care about. They can be collected on real and target hardware platforms or approximated. 

\subsubsection{Latency}
The latency is used to evaluate the running speed of a deep model making inferences. It is usually measured in the form of \textit{running time per inference} (e.g., millisecond) \cite{ioannou2017deep} or \textit{inferences per unit time} (e.g., batches/s or images/s) \cite{Ma_2018_ECCV}. In practice, the precise latency is an average value computed on a large batch or several bathes \cite{ioannou2017deep}. Some works collect this information through implementations on real devices including GPUs, TPUs, and CPUs \cite{wang2019haq,Ma_2018_ECCV,yang2020note,ioannou2017deep}. It should be noted that the reliability is unknown when evaluating a deep model's latency not on its target mobile devices (e.g., smartphone CPUs) but on non-mobile devices (e.g., GPUs). In addition to directly measuring latency, some researchers try to approximate it \cite{proxyless,syed2021generalized}. The lookup table is a latency approximation method that enumerates all the possible layers that a family of models can have along with the latency of each of the layers \cite{syed2021generalized}. It is hardware and model family-specific and requires a significant amount of time to maintain a large and dynamic database. In addition, this simple summation of the latency of an individual does not take memory access cost and parallelism into consideration, and thus shows low precision. A latency prediction model is another latency approximation method that models the network latency as a function of network structures and/or hardware parameters \cite{proxyless,wang2020gpgpu,syed2021generalized}. This approach can make latency differentiable to be directly involved in an objective function for gradient-based optimization \cite{proxyless}. 

\subsubsection{Memory Usage}
Different from latency where we care about the total inference time of a sample/batch, the \textit{peak} memory usage during inference is our major concern \cite{liberis2019neural,lin2020mcunet}. As long as the peak memory usage is lower than the memory capacity, a model is able to run on the device. It is not necessary to keep memory usage as low as possible. Extreme scenarios with maximal memory saving are considered. The peak memory usage is dominated by the intermediate tensors (so-called activation metrics), so a small model (i.e., a small number of model parameters) doesn't guarantee a low peak memory usage. For example, at similar ImageNet accuracy (70\%), even though MobileNetV2 \cite{mobilenetv2} reduces its model size by 4.6$\times$ compared to ResNet-18 \cite{resnet}, the peak memory requirement increases by 1.8$\times$ \cite{lin2020mcunet}. Thus, it is highly recommended to evaluate peak memory usage when targeting a device with quite constrained memory resources. 

In addition to the peak memory usage, which considers the on-chip memory, storage/external-memory usage is also an essential evaluation metric. It mainly restricts the model size of which model parameters occupy the largest proportion. Therefore, the \textit{bit-}precision of parameters has a crucial impact on the external-memory usage.

\subsection{Effectiveness Evaluation Metrics}
The metrics for evaluating the effectiveness of an efficient model are quite diverse and mainly dependent on the targeting tasks. Vision tasks, such as image recognition, object detection, and semantic segmentation, are such commonly used benchmark tasks \cite{Ma_2018_ECCV,mobilenetv2,tan2019mnasnet}. Different tasks have different evaluation metrics: top-1 or top 5 accuracy for image recognition \cite{tan2019mnasnet}, mAP  for object detection \cite{tan2019mnasnet}, and mIOU for semantic segmentation \cite{mobilenetv2}. In addition to vision tasks, audio tasks, such as keyword spotting, are leveraged as an evaluation task \cite{lin2020mcunet}. However, accuracy is also used as the evaluation metric in this case.

\subsection{Design Automation Cost Metrics}
As the price of freeing human efforts, design automation normally demands an excessive computational cost that prohibits its wide deployment.
\subsubsection{GPU Hours}
GPU hours/days are metrics used to evaluate the \textit{time} cost of a design automation method especially a NAS-based method \cite{zoph2018learning,he2021automl}. The GPU days can be defined as:
\begin{equation}
    \text{GPU days} = N \times t,
\end{equation}
where $N$ denotes the number of GPUs, and $t$ denotes the number of days that are used for searching \cite{liu2021survey}. GPU hours have a similar definition. 
At the early stage, it requires several or even tens of days for searching \cite{DBLP:conf/iclr/ZophL17,zoph2018learning}, while currently researchers have pushed the time to the magnitude of multiple hours \cite{dong2019searching}. 

\subsubsection{GPU Memory}
Although differentiable neural architecture search has reduced the cost of GPU hours considerably, it suffers from an intensive GPU memory cost. The consumption of GPU memory depends on the size of the candidate set for searching. Specifically, the required memory grows linearly w.r.t. the number of choices in a candidate set \cite{liu2018darts}. This issue restricts the search space size that prevents the capability of discovering novel and strong models. Some works have targeted this issue and achieved impressive progress \cite{proxyless}. Thus, it is important to involve GPU memory for a thorough evaluation. 

%% file: content/3.automated_search.tex
\section{Search for Efficient Deep Learning Models}\label{section:search}
Driven by the growing demand for mobile applications, efficient deep learning models have gained explosive attention. A tremendous number of studies have been proposed to explore manually designed efficient neural architectures or modules and achieved impressive progress \cite{yang2021condensenet,Ma_2018_ECCV,mobilenetv2,xu2020squeezesegv3,qiu2021miniseg}. Though the notable success, it is challenging for human engineers to heuristically exhaust the design space to trade off accuracy and hardware constraints. Hardware-aware neural architecture search plays an influential role in advancing this field as it automates the design process to find an optimal solution. Similar to regular NAS \cite{elsken2019neural}, the hardware-aware NAS also has three components, i.e., search space, search strategy, and performance estimation strategy, but with additional freedom or constraints (Fig. \ref{fig:flow} left bottom). In this section, we review recent achievements from these three aspects and summarize the main results in TABLE \ref{tab:search}.

\subsection{Efficient Search Space}
A search space is the basis of NAS and determines what architectures NAS can discover in principle and their performance upper-limit \cite{elsken2019neural}. A well-defined search space can not only accelerate the search process but also promote the searched model's performance \cite{zoph2018learning,liu2018hierarchical}. 
Since some manually explored efficient neural architectures have achieved considerable advances, it is an intuitive yet practical idea to construct a search space with the heuristics of these hand-crafted efficient structures, and leverage the NAS technology to automatically discover novel efficient models upon this search space. There are usually two components that define a search space: (i) \textit{operators} that each layer executes, and (ii) a \textit{backbone} that decides the topological connections of these layers.

\begin{figure*}[ht]
\centering
\subfloat[DSConv]{
    \includegraphics[width=.23\textwidth]{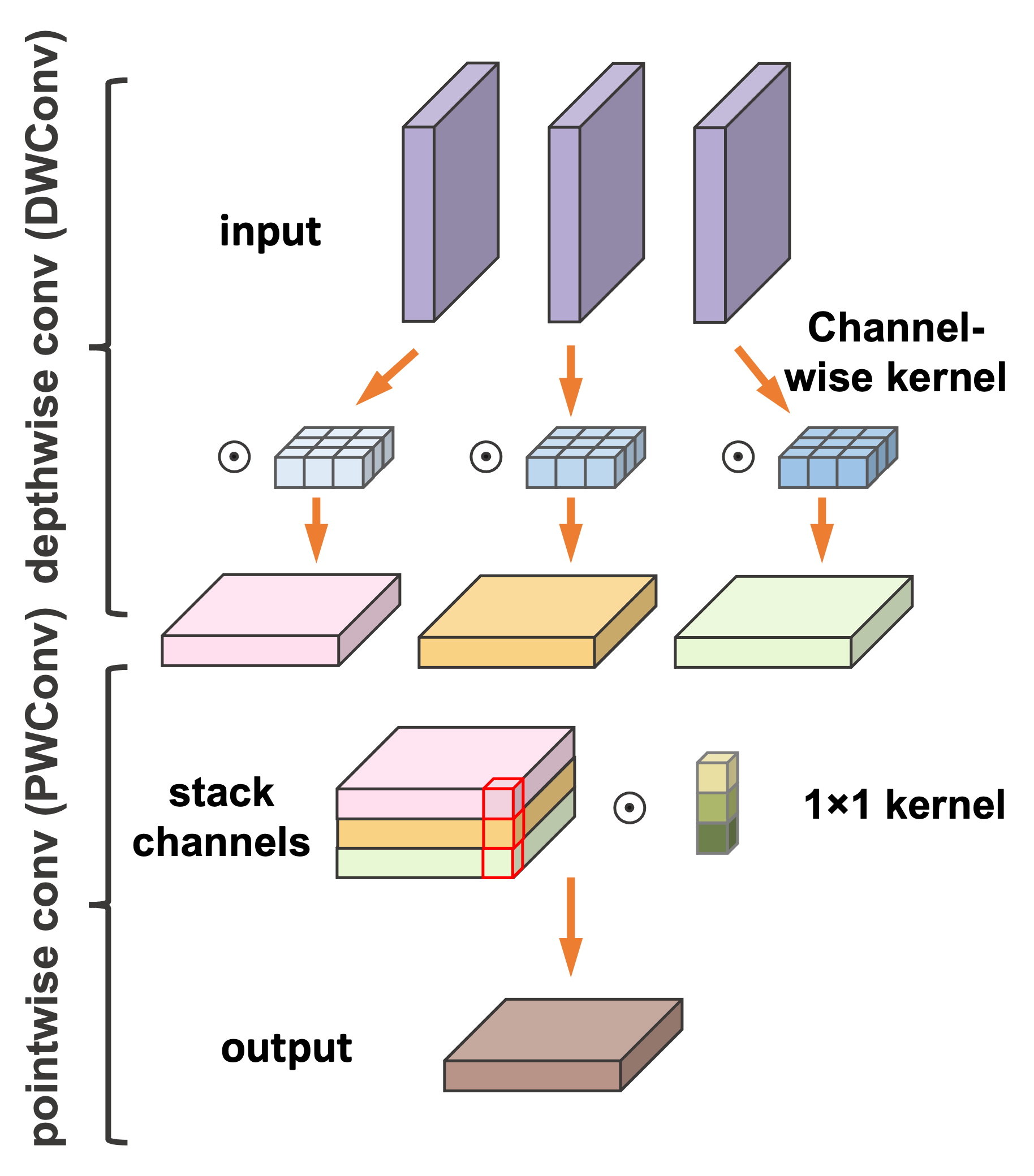}
    \label{fig:dwsconv}}
\subfloat[MBConv]{
    \includegraphics[width=.22\textwidth]{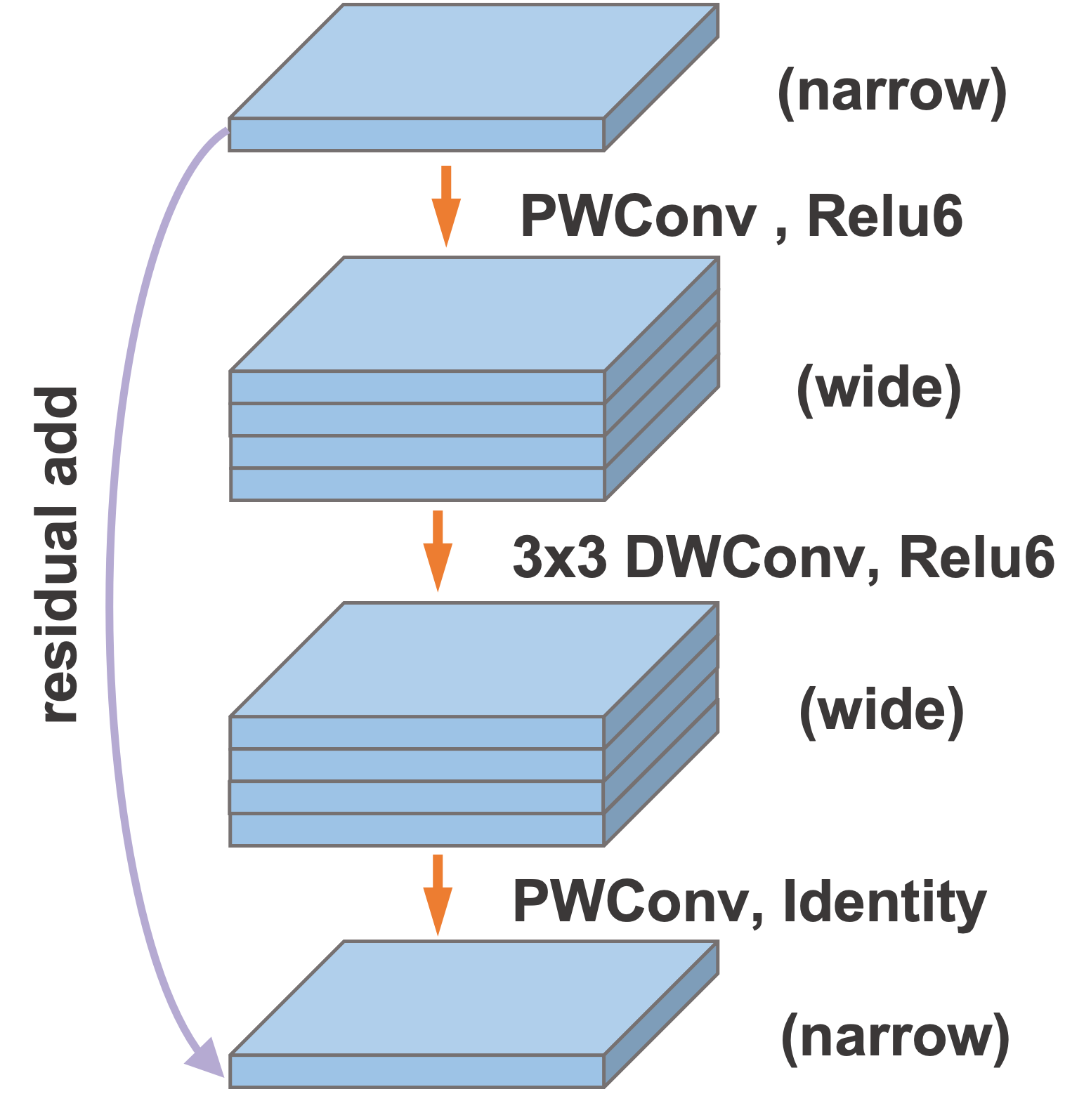}
    \label{fig:mbconv}}
\subfloat[MBConv+SE]{
    \includegraphics[width=.24\textwidth]{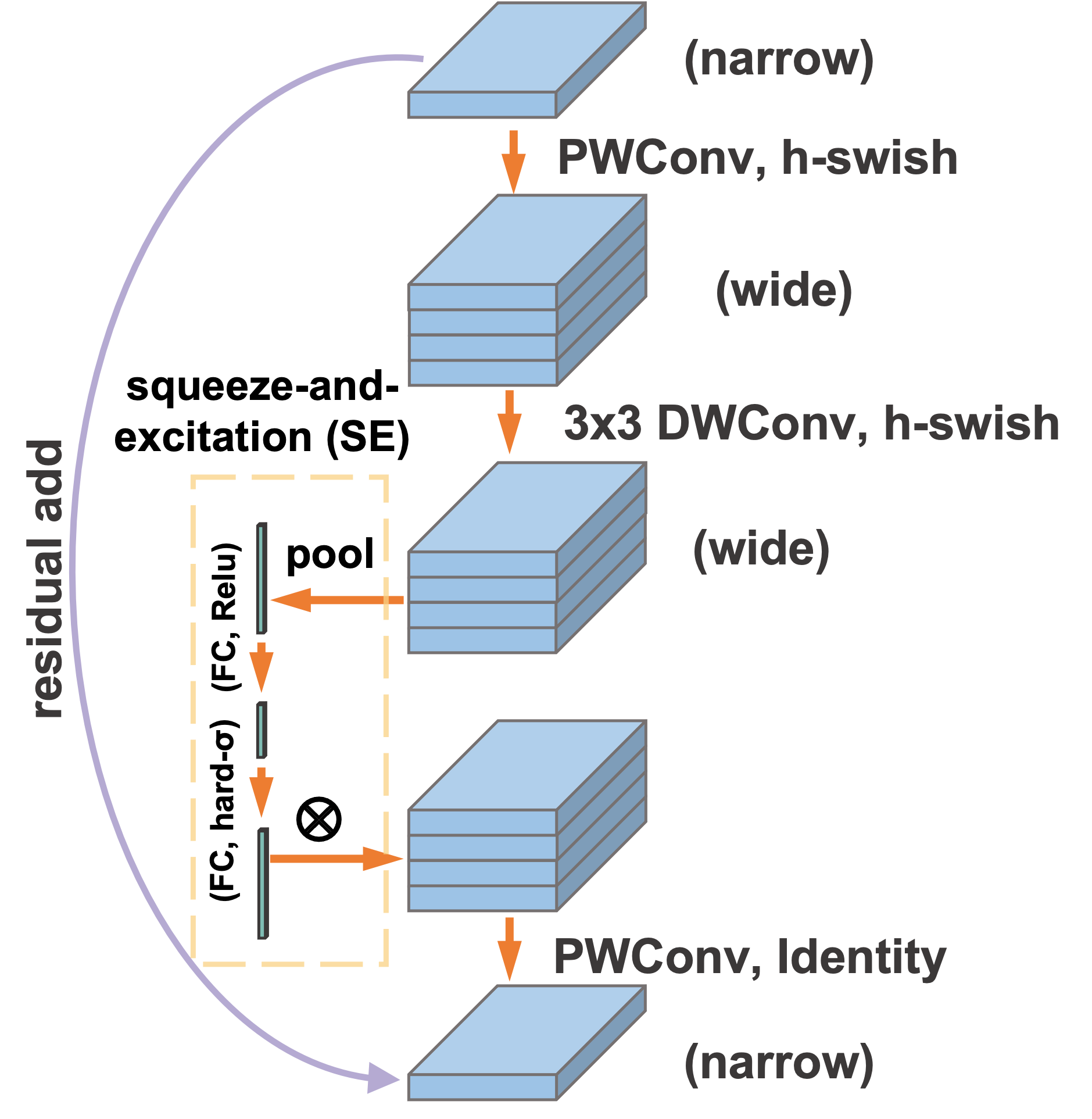}
    \label{fig:mbconv+se}}
\subfloat[GConv+channel shuffle]{
    \includegraphics[width=.28\textwidth]{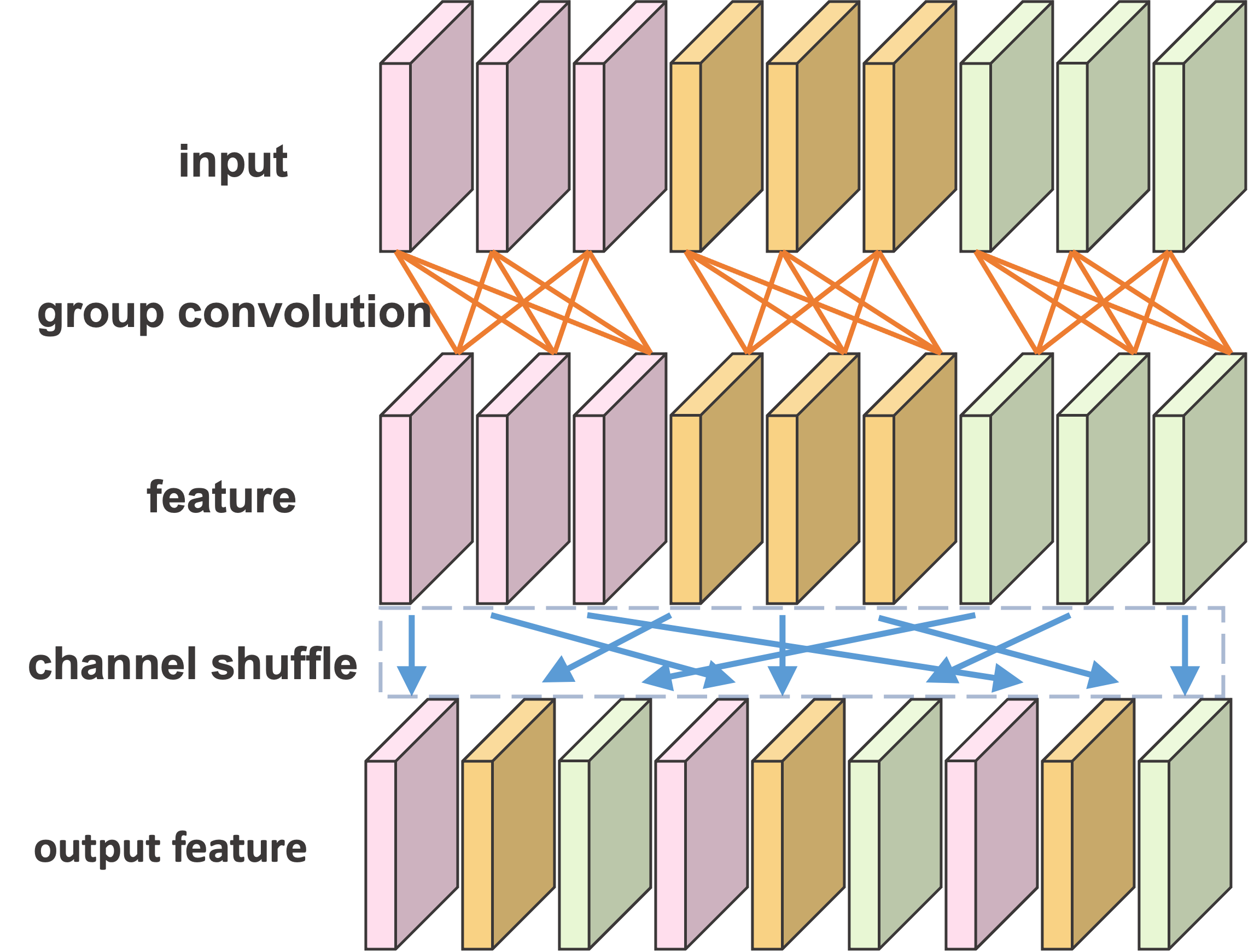}
    \label{fig:gconv}}
\caption{Schematics of efficient operators. (a) DSConv: depthwise separable convolution; (b) MBConv: mobile inverted bottleneck convolution; (c) MBConv+SE: MBConv with the squeeze and excitation module; (d) GConv+channel shuffle: group convolution with channel shuffle.}
\label{fig:operator}
\end{figure*}

\subsubsection{Operators}
Operators, such as convolutional layers or fully connected layers, are basic building components for a deep learning model. A simple way to build a deep learning model is to directly stack several operators, such as VGG Net \cite{very2015karen}. Other work connects several neural network layers to construct a motif (e.g., a residual block)
and builds a model by repeating and arbitrarily connecting these motifs \cite{densely2017huang,resnet,elsken2019neural}. In this paper, we use the term "operator" to also represent a motif, which is usually used as a whole and regarded as a basic component of an end-to-end model. 

As early-developed and widely-demonstrated neural architectures, the convolutional layer and its variants are the dominant operators of many deep learning models, especially those in the computer vision area. A \textit{standard convolution} can be denoted as:
\begin{equation}
    \text{Conv}(\textbf{W}, \bm{X})_{(i,j)}=\sum_{m,n,k}^{M,N,K}{\textbf{w}}_{(m,\;n,\;k)}\cdot \bm{x}_{(i+m,\; j+n,\;k)},
\end{equation}
where $\textbf{W}$ is the convolution kernel weight, $\bm{X}$ is the input feature maps to a convolution layer, $(i,j)$ is the coordinate of an output feature map, and $(m,n,k)$ is the coordinate of the convolutional kernel.
NSGA-Net \cite{lu2019nsga} utilizes standard convolutions as the operator and devotes to automatically determining their connections in a block-based manner. As it only seeks to optimize the connection of operators, limited efficiency is gained.  
Scheidegger et al. \cite{scheidegger2020constrained} and Lu et al. \cite{lu2019neural} also consider standard convolutions but, instead of optimizing the connections, they allow to search the convolution hyperparameters such as the filter size, the stride and the number of filters. 

In addition to the standard convolutions, extensive manually-designed variants have been demonstrated both efficient and effective. Therefore, it is intuitive to employ these architectures to construct an efficient search space. 
The \textit{depthwise separable convolution (DSConv)}, which can significantly reduce the number of parameters and computation and works as the primary module in MobileNet \cite{howard2017mobilenets}, is such a widely used efficient operator. As shown in Fig \ref{fig:dwsconv}, a DSConv block is made up of two components: a depthwise convolution and a pointwise convolution. The \textit{depthwise convolutions (DWConv)} applies a single filter per each input feature map (input depth) for spatial filtering:
\begin{equation}
    \text{DWConv}(\textbf{W}, \bm{X})_{(i,j)}=\sum_{m,n}^{M,N}{\textbf{w}}_{(m,\;n)}\cdot \bm{x}_{(i+m,\; j+n)}.
\end{equation}
The \textit{pointwise convolution (PWConv)}, a simple $1\times1$ convolution, is used to create a linear combination of the output of the depthwise convolutions for feature fusion:
\begin{equation}
    \text{PWConv}(\textbf{W}, \bm{X})_{(i,j)}= \sum_{k}^{K}{\textbf{w}}_{k}\cdot \bm{x}_{(i,\; j)}.
\end{equation}
Thus, by combining DWConv and PWConv, the DSConv is denoted as:
\begin{equation}
\begin{aligned}
    & \text{DSConv}(\textbf{W}_{p},\textbf{W}_{d},\bm{X})_{(i,j)}= \\
    & \;\;\;\;\;\;\;\;\;\; \text{PWConv}(\textbf{W}_{p},\text{DWConv}(\textbf{W}_{d}, \bm{X})_{(i,j)})_{(i,j)}.
\end{aligned}
\end{equation}
LEMONADE \cite{DBLP:conf/iclr/ElskenMH19} adopts both DSConvs and standard convolutions as the basic operators of its search space and supports increasing the number of filters and pruning filters to search efficient neural architectures. 
RENA \cite{zhou2018resource} also includes DSConvs but allows to automatically decide not only their hyperparameters but also whether they should be used in each layer. 
In addition to the standard DSConv, ProxylessNAS \cite{proxyless} includes a variant, namely dilated depthwise separable convolution, in its search space as well. The search engine can choose between the standard one and the variant. 

In MobileNetV2, a more advanced mobile convolution block, \textit{mobile inverted bottleneck convolutions (MBConv)}, is proposed \cite{mobilenetv2} and soon becomes a popular operator in favour of an efficient search space. Fig. \ref{fig:mbconv} illustrates the structure of MBConv. It is made up of three convolutions and one residual connection. 
First, a PWConv is applied to expand the input feature map to a higher-dimensional space so that non-linear activations (ReLU6) can better extract information. 
Then, a depthwise convolution is performed with $3\times3$ kernels and ReLU6 activations to achieve spatial filtering of the higher-dimensional feature maps. Furthermore, the spatially-filtered feature maps are projected back to a low-dimensional space with another pointwise convolution. Since the low-dimensional projection results in loss of information, linear activation is used after pointwise convolution. Finally, an optional residual connection (depending on whether the stride of the depthwise layer is 1) is added to combine the original input and the output of the low-dimensional projection. Note that the last two convolutions (depthwise convolution and pointwise convolution) are essentially a DSConv with dimension reduction. There are multiple works on purely using MBConv to construct an efficient search space but with different backbones or search strategies \cite{singlepath2019,fang2020,efficientnet}. Xiong et al. \cite{xiong2021mobiledets} argue that DSConv is inexpensive based on FLOPs or the number of parameters, which are not necessarily correlated with the inference efficiency, so they propose a fused MBConv layer that fuses together the first pointwise convolution and the subsequent depthwise convolution into a single standard convolution. They achieve higher mAP and lower latency on EdgeTPU and DSP than the pure MBConv search space. Li et al. \cite{li2021searching} provide an in-depth comparison between fused MBConv and MBConv and summarize that fused MBConv has higher operational intensity but higher FLOPs than MBConv depending on the shape and size of filters and activations. Therefore, they add both fused MBConv and MBConv into the search space to let the search strategy determine automatically.

In a more popular manner, MnasNet \cite{tan2019mnasnet} upgrades MBConv with a \textit{squeeze and excitation (SE)} module\cite{hu2018squeeze} after the DWConv for attentions on feature maps (as shown in Fig. \ref{fig:mbconv+se}). Specifically, the S (squeeze) procedure first converts each individual feature map into a scalar descriptor using {\it global average pooling} so the input feature maps are converted into a vector $\bm{z}\in\mathbb{R}^n$ with its $k$-th element calculated by:
\begin{equation}
    z_k=\text{GlobalArgPool}(\bm{x}_k)=\frac{1}{H\times W}\sum_{i=1}^{H}\sum_{j=1}^{W}\bm{x}_{(i,j,k)},
\end{equation}
where $\bm{x}_{(i,j)}$ is the $(i,j)$-th element of the $k$-th input feature map, which is of size $H\times W$.
The E (excitation) procedure converts the S procedure's output $\bm{z}$ into a vector of activations $\bm{s}$ using the gating mechanism of two fully connected layers:
\begin{equation}
    \bm{s}=\sigma(\textbf{W}_2\cdot\text{ReLU}(\textbf{W}_1\cdot\bm{z})),
\end{equation}
where $\sigma(\cdot)$ is the {\it sigmoid} activation function. The final output of the SE module is $\bm{\tilde{X}}$ with its $k$-th feature map denoted as:
\begin{equation}
    \bm{\tilde{x}}_k=\text{SE}(\bm{x}_k)=s_k\cdot\bm{x}_k.
\end{equation}
MobileNetV3 \cite{howard2019searching}, the latest version of the MobileNet series, also adopts the MBConv plus SE operator and further enhances it by using the hard-swish (HS) \cite{semantic2019avenash} nonlinearities instead of ReLU and replacing the sigmoid function in SE with hard sigmoid \cite{matthieu2015binary}. The authors also point out that HS in deeper layers is more beneficial. This choice is based on the fact that the sigmoid function is expensive to deploy on mobile devices. Thereafter, many more works embrace the MBConv plus SE operator in their efficient search space \cite{DBLP:conf/eccv/YuJLBKTHSPL20,chu2020moga,cai2020once,fbnetv32021dai,wan2020fbnetv2}. However, SE is not always paired with MBConv due to either not being supported by hardware \cite{chu2021discovering} or unfavourable performance \cite{mnasfpn2020chen}. Some other works maintain diverse convolution operators (i.e., standard convolution, DSConv, MBConv, and MBConv+SE) in the search space and let the search strategy choose automatically \cite{yan2021light,tan2019mnasnet}. 
The commonly used searchable parameters of the convolution operator family are kernel sizes, the number of output channels, expansion ratios (MBConv, MBConv+SE), and SE ratio (MBConv+SE). Although different works do not have exactly the same searching ranges, they are basically similar. For example, Stamoulis et al. \cite{singlepath2019} consider kernel sizes of $\{3,5\}$ and expansion ratios of $\{3,6\}$; Fang et al. \cite{fang2020} consider kernel sizes of $\{3,5,7\}$ and expansion ratios of $\{3,6\}$; Cai et al. \cite{cai2020once} consider kernel sizes of $\{3,5,7\}$ and expansion ratios of $\{3,4,6\}$.

Besides the above convolution operators of the MobileNet family, the \textit{group convolution (GConv)} and its variants \cite{xie2017aggregated,alex2012,zhang2018shufflenet} are also considered to be significant operators for constructing an efficient search space \cite{xu2021vipnas,dong2018dpp,fbnet2019wu,resource2019,hsu2018monas}. Fig. \ref{fig:gconv} illustrates the structure of GConv, where we take into account the feature map dimension of convolutional layers. The feature maps and filters are divided into $G$ ($G=3$ in Fig. \ref{fig:gconv}) groups respectively: $\bm{X}=\{\bm{X}_1, \bm{X}_2, ..., \bm{X}_G\}$ and $\textbf{W}=\{\textbf{W}_1\,\textbf{W}_2\, ..., \textbf{W}_G\}$.
In GConv, the convolution is only performed within each group so the output $ \bm{\tilde{X}}$ is denoted as:
\begin{equation}
    \bm{\tilde{X}}=\{\textbf{W}_1\otimes\bm{X}_1,\textbf{W}_2\otimes\bm{X}_2, ..., \textbf{W}_G\otimes\bm{X}_G\},
\end{equation}
where $\otimes$ is the convolution operation between two sets.
In this way, GConv can not only reduce parameters and computation but also provide a simple way to model parallelism. Therefore, AlexNet \cite{alex2012} can be trained on multiple GPUs with only 3GB RAM each. Note that depthwise convolution is a special case of GConv with the number of groups being the same as the number of channels. A channel shuffle usually comes after GConv to enable inter-group communications \cite{zhang2018shufflenet}. Xu et al. \cite{xu2021vipnas} include GConv in their search space with the number of groups as searchable from 1 (standard convolution) to N (the number of input channels). 
FBNet \cite{fbnet2019wu} and RCAS \cite{resource2019} encompasses a new convolution operator in their search space by replacing the first and last pointwise convolution in MBConv with $1\times1$ GConv. This design expands MBConv but also allows the search strategy to automatically determine whether this expansion is needed. However, their allowed maximum group amount is quite small (2 for FBNet and 4 for RCAS).
DPP-Net \cite{dong2018dpp} and MONAS \cite{hsu2018monas} also contain a variant of GConv, \textit{Learned Group Convolution (LGConv)}, which is the key operator of CondenseNet \cite{huang2018condensenet}, in their search space. The LGConv prunes away unimportant filters with low magnitude weights further reducing the computational complexity on top of the GConv \cite{xie2017aggregated}.

Some other studies use customized operators or non-convolution operators in their efficient search spaces. FasterSeg \cite{fasterseg2020chen} proposes a \textit{zoomed convolution}, where the input is sequentially processed with bilinear downsampling, standard convolution, and bilinear upsampling. The authors demonstrate that the zoomed convolution has 40\% latency trim compared to a standard convolution on a GTX 1080i GPU. Different from previous works, which focus on 2D processing, Tang et al. \cite{tang2020searching} target 3D scenes. They propose sparse point-voxel convolutions and allow to search for channel numbers and network depth. However, as the computation of 3D CNN increases more significantly with increasing kernel sizes than 2D CNN, the authors keep the kernel size as a constant of 3. HR-NAS \cite{hrnas2021ding} also involves Transformer \cite{attention2017ashish,david2019evolved} in addition to convolutions due to its recent success in computer vision \cite{nicolas2020eccv,alex2021iclr}. The authors design a lightweight Transformer that requires less computation when facing high-resolution images. Convolutional channels and Transformer queries are progressively reduced during the search. In order to facilitate high parallelism of convolution operators on TPUs and GPUs, Li et al. \cite{li2021searching} add space-to-depth/batch into the search space to increase the depth and batch dimensions. The results show that even without the lowest FLOPs, their searched EfficientNet-X models are the fastest among compared model families on TPUs and GPUs. The main reason is that EfficientNet-X models strike a balance between FLOPs (lower is good for speed) and operational intensity (higher is good for speed).

In addition to parametric operators, non-parametric operators are critical components of an efficient search space as well. \textit{Pooling} is usually coupled with convolutions in hand-crafted CNN model family, and so does in convolution-based search space \cite{resource2019,mnasfpn2020chen}. It can help to reduce redundant information (favourable for accuracy) and computation (favourable for speed) without requiring additional parameters. \textit{Skip} is another widely adopted operator that impacts the topology of achieved models. It can have two concepts: 1) \textit{drop skip} directly feeding input to output without any actual computations, i.e., the entire layer is dropped and thus the model depth is reduced \cite{singlepath2019,fbnet2019wu,fang2020}; 2) \textit{residual skip} providing an identity residual connection in parallel with another computation operator \cite{DBLP:conf/iclr/ElskenMH19,mnasfpn2020chen}. The residual connection can be achieved by either concatenation or by addition \cite{DBLP:conf/iclr/ElskenMH19}. \textit{Activation functions} are another searchable non-parametric operators that impact both accuracy and speed \cite{resource2019,li2021searching}. The computation cost of activation functions decreases when going deeper since the resolution of feature maps decreases \cite{howard2019searching}. In some research, the non-linearity is associated with convolution operators as a whole, and the operator together with its activation function is determined automatically \cite{wan2020fbnetv2}.

\subsubsection{Backbones} After defining the operators that a model can have, it is essential to determine how many operators there are and how these operators are connected, i.e., the backbone of a model. However, operators and backbones are not completely isolated, such as the skip operator, which removes layers or adds connections and can change a model's backbone. Depending on the connection topology, backbones can be roughly classified into two categories: chain-structured and multi-branch.

\begin{figure*}[ht]
\includegraphics[width=1\textwidth]{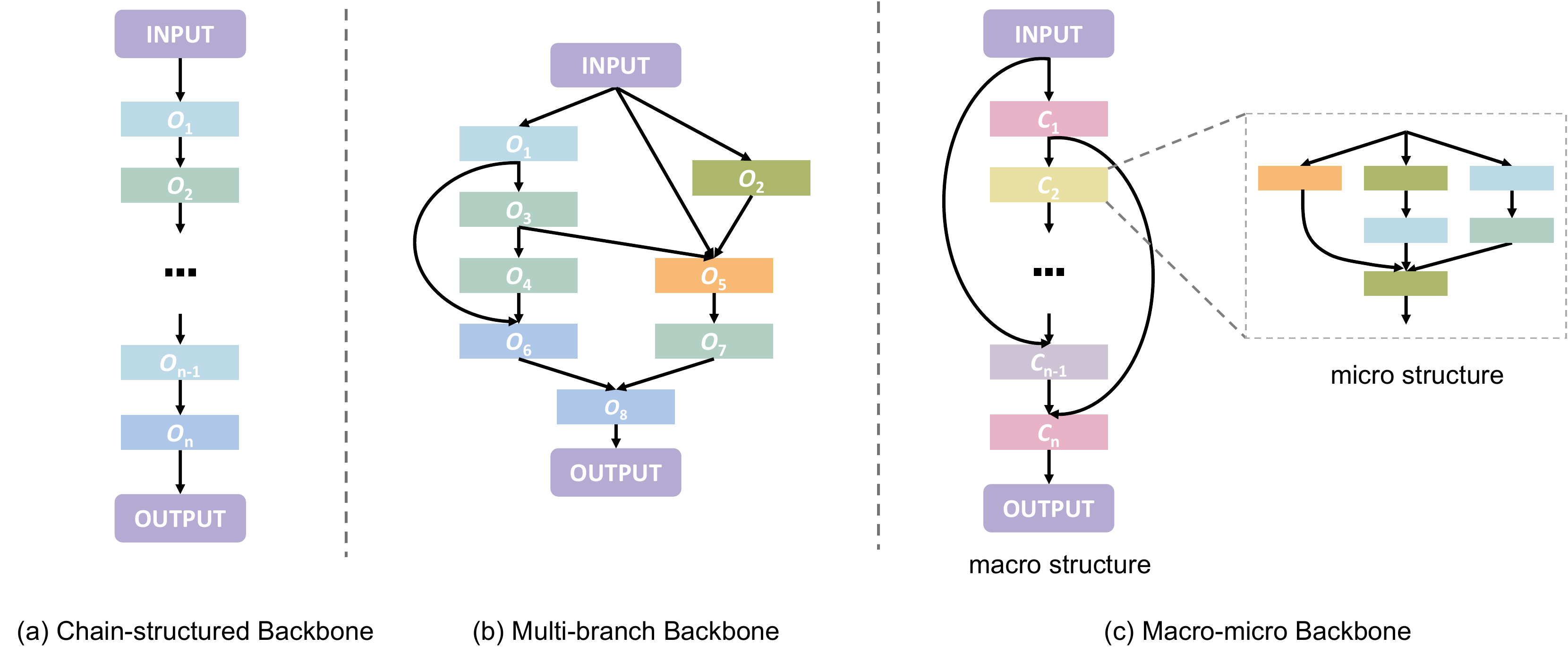}
\caption{Schematics of different backbones. Each block represents an operator $O_i$ in (a) and (b), and represents a cell $C_i$ in (c), and different colours indicate different operator/cell types. The arrows indicate the information flow direction. In the macro-micro backbone, the macro-structure describes the topology of multiple cells, each of which has a micro-structure of operators.}
\label{fig:backbone}
\end{figure*}

\textbf{Chain-structured Backbones.} The chain-structured backbone \cite{elsken2019neural} is the earliest and simplest topological structure of a neural network. It directly stacks multiple operators in sequence. As shown in Fig. \ref{fig:backbone} (a), the $i$th operator ($O_i$) takes the output of the $i$-1th operator ($O_{i-1}$) as input and its output serves as the input of the operator ($O_{i+1}$). Therefore, the topological connection of different operators is determined and not searchable. However, it is possible to search how many operators there are, e.g., via the {\it drop skip}. 
This simple backbone is a favourable practice of many hand-crafted CNN models \cite{alex2012,zhang2018shufflenet,howard2017mobilenets,mobilenetv2}. 

An intuitive implementation is to stack operators layer by layer \cite{resource2019,cai2020once}. 
Motivated by the design principle of existing success \cite{mobilenetv2,zhang2018shufflenet}, the chain-structured backbone is more commonly implemented in a cell-based, a.k.a. block- or stage-based, manner \cite{efficientnet,yan2021light,wan2020fbnetv2,tang2020searching,fbnet2019wu,fang2020}, where a backbone is composed of multiple chain-structured cells, each of which contains multiple chain-structured operators. The operators in the same cell can be either identical or diverse in hyperparameters but the same in type. In the sequence of cells, it is a usual principle that the input resolutions are reduced and widths are increased gradually. To achieve the reduced resolutions, the first or last operator in a cell usually has manually set strides and widths. This cell-based implementation is not only effective but also reduces the search space.
Wu et al. \cite{fbnet2019wu} design a chain-structured backbone consisting of four searchable cells. There are 8 candidate convolution operators with various expansion rates, kernel sizes and numbers of groups, and a drop skip operator for searching.
Different cells are separated by their input resolutions and widths, which are determined by manually set parameters. Yan et al. \cite{yan2021light} also follow the same practice to construct their backbone but with more flexibility in searchable parameters. In these studies, it is usually required to predefine some parameters of the backbone, like the number of cells. 

In comparison with setting backbone structure empirically, some work directly embraces the backbone of existing models as a starting point and then searches beyond that \cite{chu2020moga,xiong2021mobiledets,hsu2018monas,proxyless,singlepath2019,xiong2021mobiledets}. This design principle relieves experts' burden of search space design by reusing prior knowledge, which is important in NAS \cite{chen2021modulenet}. MONAS \cite{hsu2018monas} relies on the backbone of a simplified version of AlexNet \cite{alex2012} to search the convolution filter sizes and amounts. Scheidegger et al. \cite{scheidegger2020constrained} investigate the backbone of MobileNetV2 \cite{mobilenetv2} and allows to search the number of operators in each cell but all operators in a cell have the same settings except for the stride, which is used to modify the output resolution. The complexity reduction is mainly obtained by lowering the channel widths and reducing the number of topological replications. In addition to following the backbone of hand-crafted models, there is a trend to using the backbone of existing searched efficient models. 
MOGA \cite{chu2020moga} adopts MobileNetV3-large \cite{howard2019searching} as its backbone and keeps the same number and type of operators. It only searches the parameters of operators, like kernel sizes, expansion ratios for MBConv, and whether SE is enabled or not. Cai et al. \cite{cai2020once} also adopts MobileNetV3, but they additionally provide the flexibility of searching the number of operators in each cell. 

\textbf{Mutli-branch Backbones.} While the chain-structured backbone is simple, it restricts the information flow to be sequential and single-path. Current hand-crafted architectures have suggested that a \textit{multi-branch} architecture, which allows multi-path information flow and {\it residual skip} connections, works impressively better than a chain-structured architecture \cite{googlenet2015,resnet}. As illustrated in Fig. \ref{fig:backbone} (b), in the multi-branch backbone, an operator $O_i$ is allowed to accept the outputs from some of its previous operators (i.e., $O_1$, $O_2$ ... $O_{i-1}$) as the input but not necessarily takes all. This setting provides more degrees of freedom on network topology. Note that a chain-structured backbone with residual skip connections, where an operator $O_i$ must receive the output of its immediate previous operator $O_{i-1}$, is a special case of the multi-branch backbone. 

The multi-branch backbone can be achieved by using an {\it insertion} operation in the search space \cite{zhou2018resource}.
However, searching for a multi-branch backbone as a whole is time-consuming and difficult to find an optimal structure. Similar to the chain-structured backbone, the cell-based principle is also widely adopted in designing a multi-branch backbone \cite{mnasfpn2020chen,lu2019nsga,efficientnet}, where the topology within a cell has a multi-branch structure. This results in a {\it \bf Macro-micro Backbone}  (Fig. \ref{fig:backbone} (c)) \cite{tan2019mnasnet,lu2019nsga,hrnas2021ding,xu2021vipnas}. In the macro-structure, {\it residual skips} are optional to provide the flexibility of multi-path information flow; in the micro-structure, operators are connected in the multi-branch fashion with the last operator (either parametric or non-parametric) assembling a single output of the cell. For example, MnasNet \cite{tan2019mnasnet} predefines a backbone of 7 sequentially stacked cells and each cell contains sequentially stacked operators with a searchable amount, types, and parameters. It further allows searchable residual skip connections among operators within a cell. This multi-branch backbone simply augments the chain-structured backbone with residual skips, resulting in more flexibility of multi-path information flow but marginal research space expansion. 
NSGA-Net \cite{lu2019nsga} and HR-NAS \cite{hrnas2021ding}, in contrast, allow more general multi-branch connections within a cell. NSGA-Net supports searchable topology while HR-NAS sets fixed multi-branch connections and searches for discarded convolutional channels and transformer queries.

Different from searching for a customized structure, following existing success is also a favoured choice in designing multi-branch backbones \cite{li2021searching,chu2021discovering,mnasfpn2020chen,howard2019searching}. Li et al. \cite{li2021searching} adopt the multi-branch backbone of the EfficientNet and only search operators. MnasFPN \cite{mnasfpn2020chen} constructs its search space based on the NAS-FPN(Lite) backbone and searches both multi-branch structures and operators in a cell for merging various resolutions. However, to reduce the search burden, it does not allow general connectivity patterns and applies limited merging connections. MobileNetV3 \cite{howard2017mobilenets} uses the backbone of MnasNet \cite{tan2019mnasnet} as the seed and proceeds layer-wise search on it. Scheidegger et al. \cite{scheidegger2020constrained} investigate the backbone of several existing models (DenseNet121 \cite{densely2017huang}, MobileNetV2 \cite{mobilenetv2}, GoogLeNet \cite{googlenet}, PNASNet \cite{DBLP:conf/eccv/LiuZNSHLFYHM18}, and ResNeXt \cite{xie2017aggregated}) and demonstrate that their approach is able to provide improved accuracy with hardware constraints and different backbones. 
While considering that searching both macro and micro structures overburdens the searching process, most works utilize the multi-branch macro backbone of existing models and only search the micro-structures \cite{proxyless,dong2018dpp,hsu2018monas}. ProxylessNAS \cite{proxyless} accepts the backbone of the residual PyramidNet \cite{pyramidnet}, which has a residual skip connection every two operators, and replaces the original operators with their own tree-structured cells \cite{cai2018path}.
DPP-Net \cite{dong2018dpp} selects the backbone of CondenseNet \cite{huang2018condensenet}, which repeats an identical cell abundant times with both residual skip and chain connections, and only searches the operators in the cell. MONAS \cite{hsu2018monas} also reuses the backbone of CondenseNet but it uses the same cell structure and searches the number of stages and growth rate.

Although the macro-micro backbone provides the highest flexibility and complexity among the above three categories, there is no evidence that it is the best choice. However, we note that no matter in which backbone category, cell-based implementation is the most common practice. This is due to the following considerations: i) an effective network usually has gradually shrinking resolutions as going deep, and a cell can have multiple operators on the same resolution to strengthen feature extraction; ii) previous experience with the manually designed network indicates that the cell-based structure is effective for deep learning; iii) the cell-based backbone provides high search efficiency by limiting the search space (e.g., the whole network can stack the same cell and a cell can have the same operator); iv) the cell-based backbone simplifies the network topology with inter- and intra-cell connectivity instead of random edges among all operators.
Some recent works reveal that search performance may be impeded by these search space design biases \cite{exploring2019}. Nevertheless, the choice of the search space principally regulates the difficulty of the search process. Current NAS algorithms are imperfect so more freedoms in the search space do not always lead to better resultant models but inversely overburden the search algorithms. With the continued improvement of NAS algorithms, it is essential to reduce design biases and construct a more general search space. 
 
\subsection{Search Strategy}\label{searchstrategy}
The search strategy describes how to explore a search space to find an optimal efficient network. 
Essentially, the search process is to find the top candidate networks regarding some evaluation metrics (e.g., accuracy and latency). However, it is computationally prohibitive to achieve these metrics of all candidate networks since it requires fully training a network to obtain its metrics. Therefore, the aim of a search strategy is to efficiently find top-ranking networks without exhaustively examining all candidate networks. 
In this section, we first conclude the \textit{search algorithms} that describe how the search proceeds and then summarize how the hardware constraint is incorporated into the search process. 

\begin{figure*}[ht]
\tiny
\centering
\subfloat[\footnotesize Bayesian optimization]{
    \includegraphics[width=.25\textwidth]{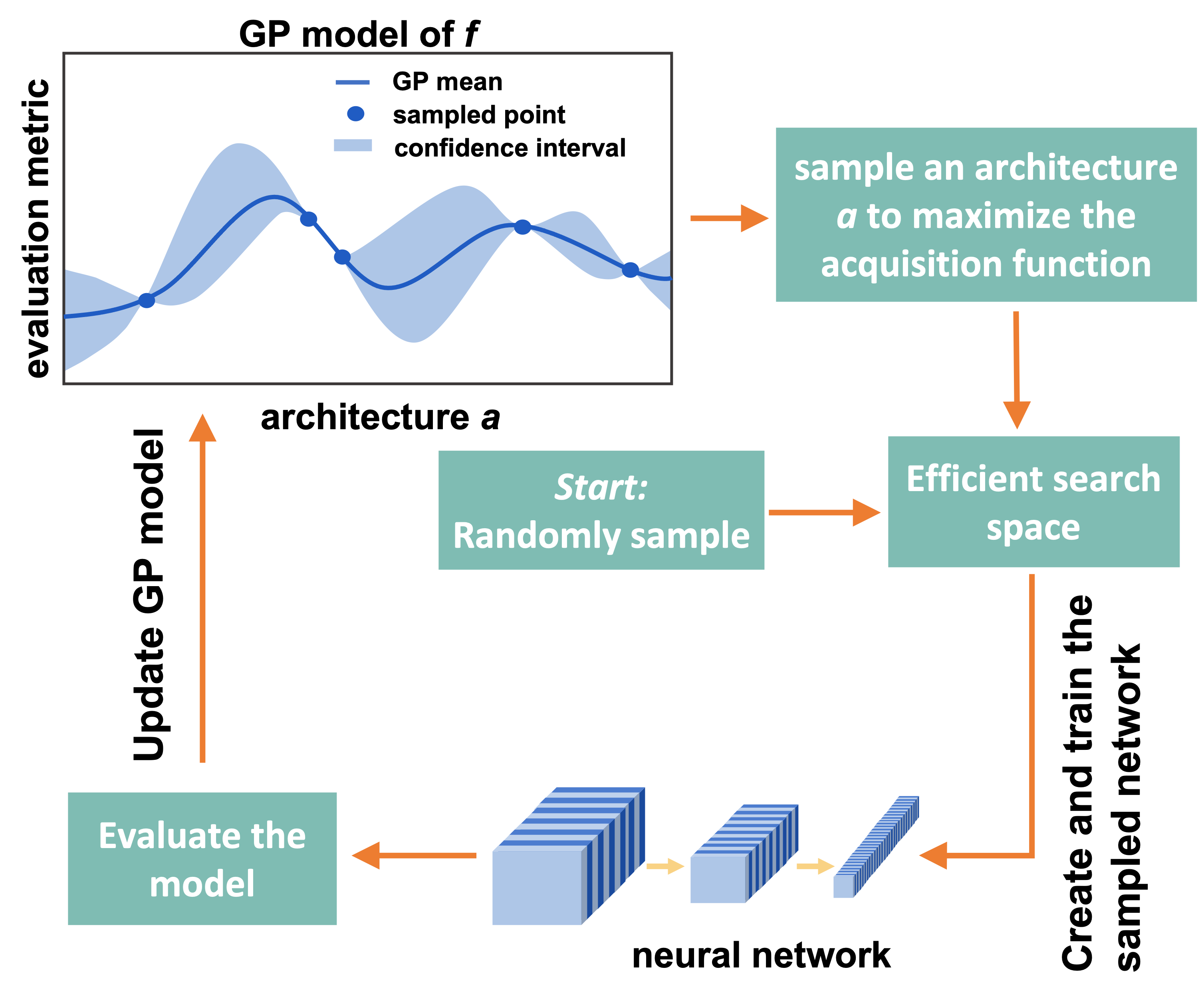}
    \label{fig:BO}}
\subfloat[\footnotesize Evolutionary search]{
    \includegraphics[width=.25\textwidth]{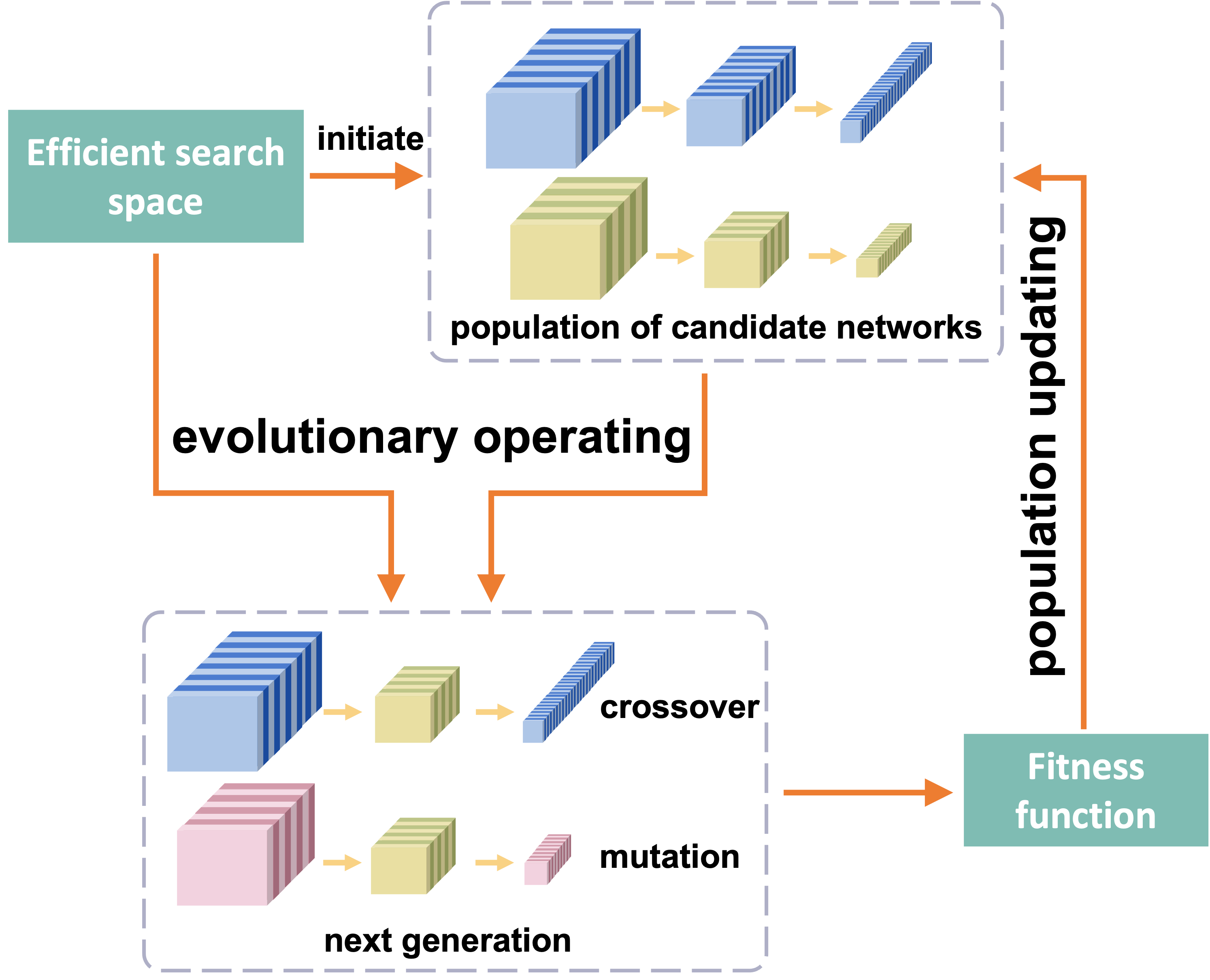}
    \label{fig:EA}}   
\subfloat[\footnotesize Reinforcement learning]{
    \includegraphics[width=.25\textwidth]{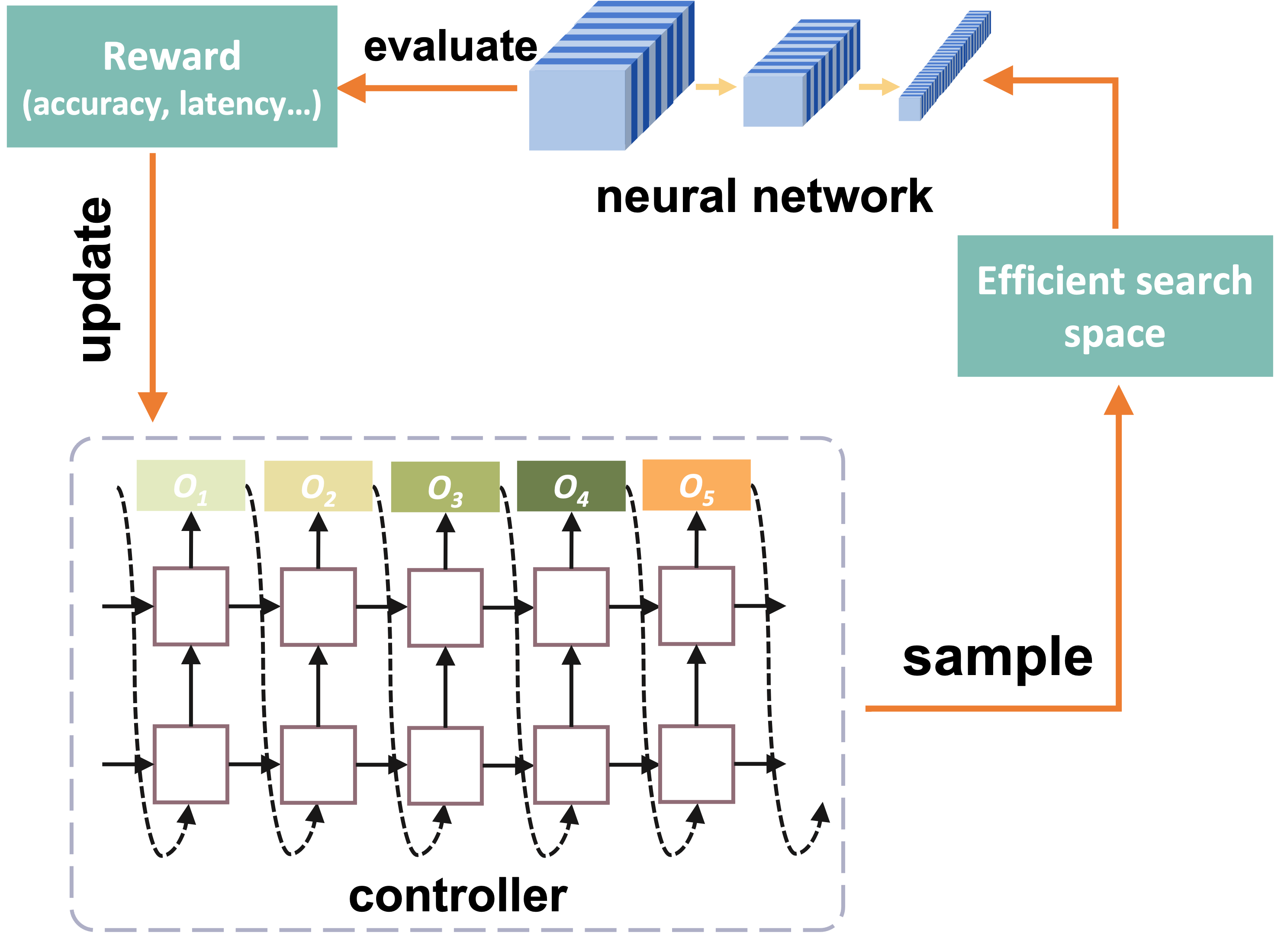}
    \label{fig:RL}}   
\subfloat[\footnotesize Differentiable search]{
    \includegraphics[width=.20\textwidth]{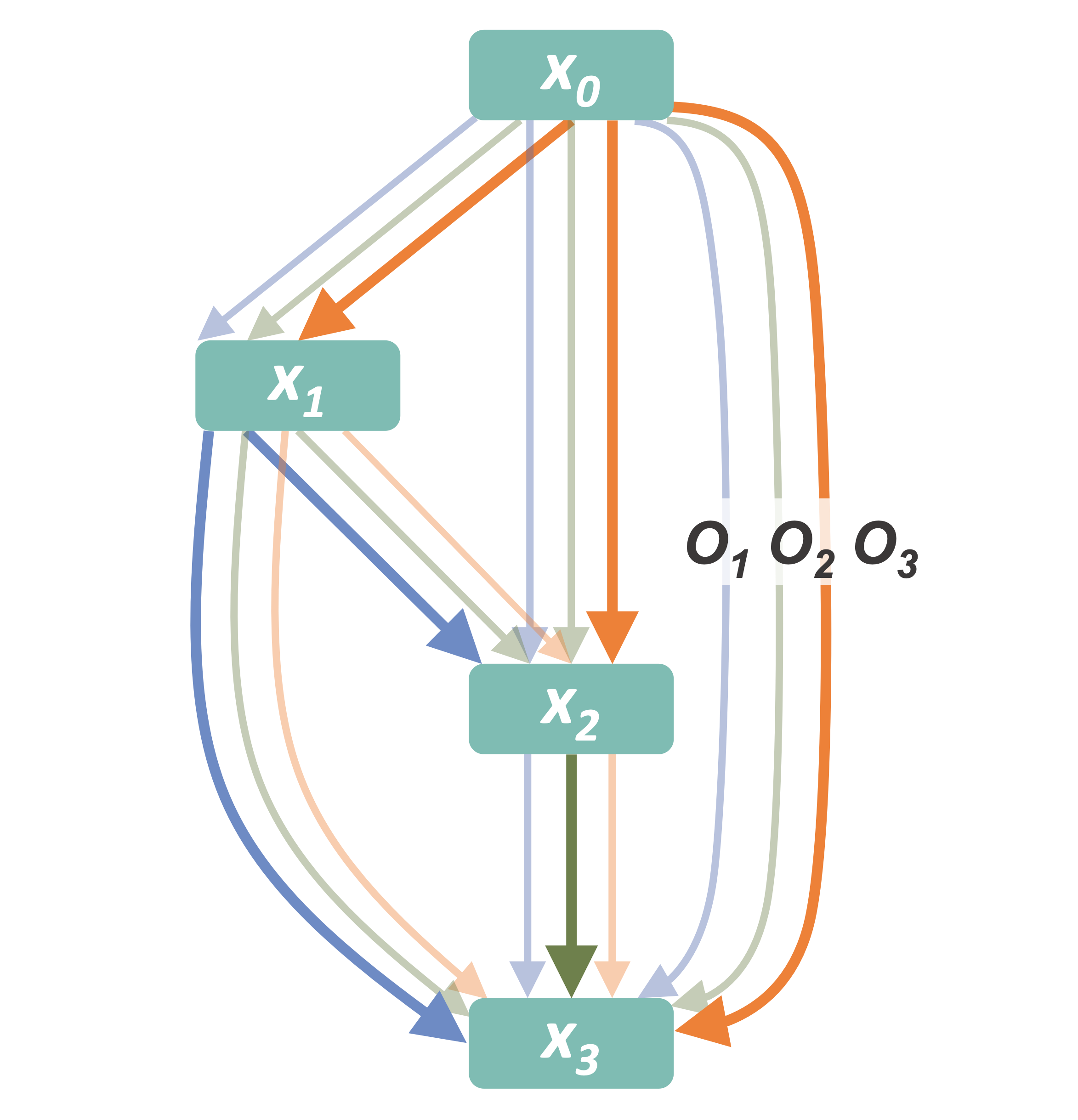}
    \label{fig:diff}} 
\caption{Schematics of different search algorithms. (a) Bayesian optimization; (b) Evolutionary search; (c) Reinforcement learning; (d) Differentiable search.}
\label{fig:algorithm}
\end{figure*}

\subsubsection{Search Algorithms}
The most intuitive and easiest search algorithm is the \textit{grid search}, which exhaustively explores the search space and evaluates every possible architecture. This approach works well for a small space \cite{efficientnet} but is terribly inefficient for a large space due to the exponentially increased number of evaluations. Another problem with this type of method is that it only supports a bounded and discrete space and needs careful selection of the grid interval. \textit{Random search} \cite{DBLP:journals/jmlr/BergstraB12}, on the other hand, samples neural architectures randomly from the search space. It can be used not only for a discrete space but also for a continuous space with a predefined distribution. This algorithm is superior to the grid search also when the search dimensions (e.g., kernel sizes, expansion ratios, and network depths) have different effects on the final performance. Random search and its simple variants are usually used when the performance of a candidate model is easy to obtain \cite{xu2021vipnas,DBLP:conf/eccv/YuJLBKTHSPL20}. Advanced search algorithms, like reinforcement learning and evolutionary search, are widely demonstrated better performance than random search \cite{zoph2018learning,DBLP:conf/eccv/GuoZMHLWS20,DBLP:conf/cvpr/BenderLCCCKL20}. In the following, we summarize the advanced search algorithms,
including Bayesian optimization, evolutionary search, reinforcement learning, and differentiable methods, regarding design automation for efficient deep learning models.  

\textbf{Bayesian Optimization (BayesOpt)} is an important approach for automated hyperparameter tuning and architecture search. Given a search space $\mathbb{S}$ that contains a large set of neural architectures, and a \textit{black-box} objective function $f(\cdot)$ from an neural architecture to an evaluation metric (e.g., accuracy), the goal of BayesOpt is to find an architecture $a\in\mathbb{S}$ that maximizes $f(\textbf{a})$: $\textbf{a}^\ast=\argmax_{\textbf{a}\in\mathbb{S}} f(\textbf{a})$. However, it is expensive to achieve $f(\textbf{a})$ because it requires fully training the architecture $\textbf{a}$ from scratch. Therefore, a statistical model, which is invariably a \textit{Gaussian process (GP)}, is used as a surrogate model. The more neural architectures (i.e., points of $f(\cdot)$) are evaluated, the less uncertainty the surrogate model has. Another important component of BayesOpt is the acquisition function for deciding which architectures are sampled and then evaluated at each iteration, which is often \textit{expected improvement (EI)}:
\begin{equation}
\begin{aligned}
   & \text{EI}(\textbf{a})=\mathbb{E}_{max}(f(\textbf{a})-f(\textbf{a}^{+}),\;0), \\
   & \hspace{3cm} \textbf{a}^{+}=\argmax_{\textbf{a}_i\in\textbf{a}_{1:t}}f(\textbf{a}_i),
\end{aligned}
\end{equation}
where $\textbf{a}$ is a sampling architecture and $f(\textbf{a}^{+})$ is the evaluation metric of the best architecture $\textbf{a}^{+}$ so far, that is until $t$-th exploration.
The general process of BayesOpt is shown in Fig. \ref{fig:BO}, where multiple architectures are first randomly sampled from the efficient search space; the sampled architectures are then trained and evaluated for updating a GP model; the acquisition function is computed based on the updated GP model to decide which neural architectures should be sampled next; the process proceeds iteratively until certain conditions are met. The best architecture among all sampled architectures is usually output as the final model. 

When additionally considering the hardware constraints, Multi-Objective Bayesian Optimization (MOBO) is commonly used \cite{DBLP:conf/dac/OdemaRDF21,DBLP:journals/sensors/YangZLLWWZ21,parsa2020bayesian,eriksson2021latency}, where GP is fitted for each objective independently, and \textit{Pareto-frontier} is identified as the set of optimal trade-off solutions of the multiple objectives. Different MOBO approaches mainly differ in how to achieve the Pareto-frontier during the acquisition process. For example, Parsa et al. \cite{parsa2020bayesian} use a Gaussian distribution to estimate the Pareto-frontier function; Eriksson et al. \cite{eriksson2021latency} use the Noisy Expected Hypervolume Improvement (NEHVI) acquisition function, which is a noise-tolerant and extended version of EI for the multi-objective setting, to sample intermediate Pareto-frontiers. 
Although BayesOpt has shown promising performance, especially in hyperparameter optimization, it is computationally expensive and struggling when handling a high-dimensional search space and multiple objectives \cite{gaudrie2019high,daulton2021multi}.

\textbf{Evolutionary Search (ES)} is a long-thriving NAS approach, which is based on the concept of biological evolution. It manages to find an optimal solution by iteratively improving upon a population of candidate solutions according to a fitness function. As illustrated in Fig. \ref{fig:EA} (the schematic of the basic ES process), a population of candidate networks is first initialized by randomly sampled from the efficient search space; then evolutionary operations, such as crossover (combination of two parents) and mutation (arbitrarily mutating operators with new ones from the search space), are applied to current population to generate next generation; lastly, all offsprings need to be tested against a fitness function, which is to select the most powerful candidate networks and update the population; the process proceeds repeatedly until stopping criteria are met. The best network in the last population is the final achieved model. The major benefit of evolutionary search is its flexibility in directly controlling the offspring generation process and population updating process \cite{DBLP:conf/eccv/GuoZMHLWS20}.

The hardware constraints are usually considered in the fitness function of ES in two ways: hard-constraint and soft-constraint, which we will present concretely in the flowing section \ref{hc}. Thus, most studies directly use existing evolutionary algorithms with customised fitness function to incorporate hardware constraints \cite{fbnetv32021dai,tang2020searching,yan2021light,DBLP:conf/iccv/MoonsNSMMLB21,cai2020once,DBLP:conf/date/LuoLHL21,DBLP:conf/eccv/GuoZMHLWS20,DBLP:conf/cvpr/DaiZWYSWDHWJVUJ19}. For example, FBNetV3 \cite{fbnetv32021dai} uses the \textit{adaptive genetic algorithms} \cite{DBLP:journals/tsmc/SrinivasP94} to realize adaptive probabilities of crossover and mutation, and customizes the fitness function with its proposed \textit{accuracy predictor} and hard hardware constraints; OFA \cite{cai2020once} adopts the \textit{regularized evolutionary search} \cite{DBLP:conf/aaai/RealAHL19}, which introduces an age property to favor younger generation during search; DONNA \cite{DBLP:conf/iccv/MoonsNSMMLB21}, FairNAS \cite{DBLP:conf/iccv/Chu0X21}, and MOGA \cite{chu2020moga} appeal to the famous NSGA-\rom{2} \cite{DBLP:journals/tec/DebAPM02}, which is a multi-objective evolutionary algorithm, to find the Pareto-optimal solution instead of the solution under hard constraints in previous studies. In contrast to directly using off-the-shelf ES algorithms, some researchers develop ES algorithms specific to the hardware efficient application \cite{DBLP:conf/iclr/ElskenMH19,lu2019nsga,scheidegger2020constrained}. LEMONADE \cite{DBLP:conf/iclr/ElskenMH19} is such a representative work, which handles various objectives differently considering that different objectives have different evaluation costs. Specifically, the efficiency objective (e.g., FLOPs) is cheap to evaluate while the accuracy objective is much more expensive as it requires fully training the model at each iteration. The authors propose to first select the architectures that fulfil the Pareto front for the cheap objectives and then only train and evaluate these networks. As traditional ES can only sample a limited amount of networks that satisfy hardware constraints, Scheidegger et al. \cite{scheidegger2020constrained} use ES algorithms to search sampling laws that can better cover the sub search space under specific constraints. Thus, better Pareto frontiers can be achieved. NSAG-Net \cite{lu2019nsga} designs an additional exploitation step after the traditional evolutionary operating to learn and leverage the history of evaluated populations. Since the evaluated populations at each iteration rank relatively high regarding the fitness score, it is quite possible that the optimal model has similarities to the evaluated populations and thus extracting common patterns from the evaluated populations can accelerate convergence. 
Although the above studies manage to lighten the search process from the algorithm perspective, it is still costly due to the requirement of training each network of the new generation at each interaction.

\textbf{Reinforcement Learning (RL)} has achieved notable success and grabbed great attention in the NAS community since Zoph and Le's work in 2017 \cite{DBLP:conf/iclr/ZophL17}. The general framework of RL for NAS is illustrated in Fig. \ref{fig:RL}: the controller, usually an RNN network, is the core component, which generates actions to sample operators or hyperparameters from the efficient search space; a neural network will then be constructed using the sampled options and trained and evaluated to achieve a reward, which will be used to update the controller using a policy gradient method \cite{DBLP:journals/ml/Williams92}. This process will be performed repeatedly until stopping criteria are met. 

The hardware information is often considered in the RL reward function. Some studies thus directly use the existing RL-based NAS approach \cite{DBLP:conf/iclr/ZophL17}, but optimize with a customized reward function \cite{hsu2018monas,zhou2018resource}. Furthermore, most recent work mainly follows the advanced RL search algorithms in two representative studies, MnasNet \cite{tan2019mnasnet} and TuNAS \cite{DBLP:conf/cvpr/BenderLCCCKL20}. Specifically, \cite{mnasfpn2020chen,howard2019searching} follow MnasNet, and \cite{xiong2021mobiledets,chu2021discovering} follow TuNAS. Similar to \cite{DBLP:conf/iclr/ZophL17}, MnasNet \cite{tan2019mnasnet} also uses RNN as the controller while maximizing the expected reward using an advanced policy gradient method, Proximal Policy Optimization (PPO) \cite{schulman2017proximal}, to alleviate the high gradient variance of the vanilla policy gradient method. TuNAS is based on ENAS \cite{pham2018efficient} and ProxylessNAS \cite{proxyless}. It improves with {\it warmup} and {\it channel masking} techniques for better search robustness and scalability. Instead of building one candidate network at each training step, TuNAS encompasses all candidate networks into a {\it supernet}. Each path of the supernet represents a candidate network. In addition, TuNAS is not a sequential decision-making process and does not rely on an RNN controller; alternatively, it uses learnable probability distribution that spans over architectural choices as the RL controller. At each step, a candidate network is sampled from the distribution, and then the portion of the supernet correlated with the sampled network is trained; a reward is calculated with the sampled network to update the probability distribution i.e., the RL controller. 

\textbf{Differentiable Search} is the most recently developed search paradigm that finds an optimal network by gradient descent \cite{liu2018darts}. As shown in Figure \ref{fig:diff}, a network is as a directed acyclic graph consisting of an ordered sequence of $N$ nodes ($N=4$ in Figure \ref{fig:diff}: $X_0, X_1, X_2, X_3$). Each node $X_i$ is a latent representation and each directed edge represents a candidate operator $O_i$ that is applied to $X_i$. For example, in Figure \ref{fig:diff}, there are three candidate operators (i.e., $O_1$, $O_2$, and $O_3$) between node $X_0$ and $X_3$. In this way, the search process is formulated as an optimal path-finding problem. All outputs of the candidate operators from a node's predecessors are summed with weights $\alpha_i$ to achieve the node representation: $X_{j}=\sum_{i=1}^{m}\alpha_iO_i(X_i)$, subject to $\alpha_i\geq0, \sum_{i=1}^{m}\alpha_i=1$. In Figure \ref{fig:diff}, edges with darker colours indicate larger weights. The weights $\alpha_i$ can be represented with a softmax function of architectural parameters $\beta_i$: $\alpha_i=\frac{exp(\beta_i)}{\sum_{i=1}^{m}exp(\beta_i)}$. The network weights ($w$) and architectural parameters ($\beta_i$) are trained alternatively with training data and validation data, respectively. This induces a {\it bi-level} optimization problem \cite{liu2018darts}:
\begin{equation}\label{eq:bilevel}
    \min_\alpha\min_{w_\alpha}\mathbfcal{L}(\alpha,w_\alpha).
\end{equation}
Therefore, unlike ES and RL, a differentiable search unifies model training and search into a joint procedure. As the whole network contains multiple parallel operators in each layer, it is called a supernet, and its subnet with one operator in each layer is a candidate network. The candidate operator with the highest associated weight is chosen to construct the final model. The training of the supernet is also the search process. 

The hardware information can be considered in the loss function and optimized when training the architectural parameters. Various differentiable search algorithms are proposed to achieve better search efficiency and performance \cite{wan2020fbnetv2,fbnet2019wu,proxyless,singlepath2019,hrnas2021ding,fang2020}. FBNetV1 \cite{fbnet2019wu} replaces the softmax function with the \textit{Gumbel softmax function} \cite{DBLP:conf/iclr/JangGP17} to better represents the subnet sampling process, and thus reduce the performance gap between the supernet and subnet. HR-NAS \cite{hrnas2021ding} progressively discards the paths with low path weights during the search to increase the search efficiency. DenseNAS \cite{fang2020} splits the search procedure into two stages: the first stage optimizes the model weights only for enough epochs and the second stage alternatively optimizes the model weights and architectural parameters. This strategy alleviates the bias of fast convergence operators. Considering the small size of the search space of the conventional differentiable search due to its requirement of loading the whole supernet into memory, many studies propose novel differentiable search algorithms to reduce the memory footprint \cite{proxyless,singlepath2019,wan2020fbnetv2,fang2020}. ProxylessNAS \cite{proxyless} proposes to factorize the task of training of all paths into training two sampled paths, which have the highest sampling probabilities, and discard the other paths temporarily at each iteration. The architectural parameters of the two selected paths are updated during training and then rescaled after training to keep the path weights of unsampled paths unchanged. In this way, the memory requirement is decreased to the level of training two subnets. Different from previous papers, which use multi-path supernets and thereby have memory issues, Single-Path NAS \cite{singlepath2019} and FBNetV2 \cite{wan2020fbnetv2} develop masking mechanisms that train a supernet with the largest hyperparameters (e.g., $7\times7$ kernel) and a mask/indicator function that determines whether to just use a small part (e.g., $5\times5$ or $3\times3$ kernel) of the largest hyperparameters. 

\textbf{Other} types of search algorithms are also reported \cite{resource2019,dong2018dpp}. Xiong et al. \cite{resource2019} propose a modified \textit{Cost-Effective Greedy} algorithm for the submodular NAS process, where starting from an empty network, each block is filled iteratively with the highest marginal gain ratio regarding both accuracy and cost. DPP-Net \cite{dong2018dpp} follows the \textit{progressive search} algorithm in \cite{DBLP:conf/eccv/LiuZNSHLFYHM18} to find the optimal operator layer by layer. Different from searching for a new model for each new resource budget, another way is to build a baseline model first and then scale it to obtain a family of models for various budgets \cite{efficientnet,li2021searching}. Specifically, they adopt the conventional RL-based hardware-aware search algorithm \cite{tan2019mnasnet} to search for a small baseline model, and use a simple grid search to determine the best scaling coefficient (i.e., $\alpha$, $\beta$, $\gamma$) for network depth, width, and input size. A family of specialized models for different budgets can be obtained by scaling up the baseline model by $\alpha^N$, $\beta^N$, and $\gamma^N$ with $2^N$ times increased resource. 

\subsubsection{Hardware-constraint Incorporation Strategy}\label{hc}
Despite some studies only using an efficient search space without considering hardware constraint \cite{zoph2018learning}, there are two strategies commonly used to consider hardware budget for finding optimal compact models. The first considers a specific hardware platform and treats its resources as \textit{hard constraint} to build a specialized model that is the most accurate under the fixed constraint.
In contrast, the second strategy is to directly search networks without considering \textit{specific} hardware constraint. This strategy, dubbed \textit{soft-constraint incorporation}, treats the model efficiency as an additional optimization objective and tries to find the Pareto frontiers. 
In this section, we summarize the design automation techniques from the above two perspectives.

\textbf{Hard-constraint Incorporation} can be easily adopted by evolutionary search \cite{yan2021light,tang2020searching,cai2020once,fbnetv32021dai,scheidegger2020constrained} since the offspring generation and selection step can be directly controlled. The incorporation process is quite straightforward: when a set of candidate networks is generated (e.g., by evolutionary operating), their hardware costs are then tested, and the networks that do not meet the constraint are simply discarded. This strategy is also easy to implement for other search algorithms, which generate a set of candidate networks one time or at each iteration, such as random search \cite{xu2021vipnas,DBLP:conf/eccv/YuJLBKTHSPL20}, grid search \cite{efficientnet}, and greedy search \cite{resource2019}.
The RL-based search algorithms can also embrace the hard-constraint incorporation strategy but in a different way \cite{zhou2018resource,tan2019mnasnet,hsu2018monas}. The general idea is to have only the accuracy in the reward function when the constraint is met; otherwise, both accuracy and the hardware objective are considered in the reward function. A typical reward formula $\mathbfcal{R}(\cdot)$ is proposed in MnasNet \cite{tan2019mnasnet}:
\begin{equation}
\mathbfcal{R}(a)=
\begin{cases}
ACC(a), & \text{if} \;\; C(a)\leq C_0\\
ACC(a)\times(C(a)/C_0)^\beta, & \text{if} \;\; C(a)>C_0
\end{cases}
\end{equation}
where $ACC(a)$ and $C(a)$ denote the accuracy and hardware cost (e.g., latency in \cite{tan2019mnasnet}) of model $a$, respectively, and $C_0$ denotes the target cost constraint. $\beta<0$ is the only tunable hyperparameter that controls the convergence speed. In \cite{DBLP:conf/cvpr/BenderLCCCKL20}, the authors empirically find that if $\beta$ is too large, the RL-controller will prefer to sample the architectures whose cost is significantly smaller than the target cost constraint. This will result in suboptimal models regarding accuracy.

\textbf{Soft-constraint Incorporation} does not target any specific constraint and is realized by adding an additional optimization objective to the accuracy objective. Since these objectives are competing, no unique optimal solution exists in the multi-objective space. Thus, Pareto frontiers are sought, especially by multi-objective BayesOpt and multi-objective evolutionary algorithms, and one specific model can be identified according to different application requirements \cite{lu2019nsga,parsa2020bayesian,DBLP:conf/iccv/MoonsNSMMLB21,chu2020moga,DBLP:conf/dac/OdemaRDF21,DBLP:conf/iclr/ElskenMH19,DBLP:journals/sensors/YangZLLWWZ21,DBLP:conf/cvpr/DaiZWYSWDHWJVUJ19,DBLP:conf/iccv/Chu0X21}. For example, ChamNet \cite{DBLP:conf/cvpr/DaiZWYSWDHWJVUJ19} designs a multi-objective fitness function:
\begin{equation}
    \mathbfcal{F}(a)=ACC(a)-[\alpha H(C(a)-C_0)]^\beta,
\end{equation}
where $C(a)$ and $C_0$ denote the hardware cost of model $a$ and the target constraint, respectively, $H$ is the Heaviside step function, and $\alpha$ and $\beta$ are positive constants. The searching objective is to maximize the fitness function: $a^*=\argmax_a(\mathbfcal{F}(a))$. This fitness function guides the search process to find a model approaching the hardware target but without a hard guarantee.
Some work also pursues the soft-constraint objective under a hard constraint with a two-phase approach, which first filters out models that do not satisfy the hard constraint and then performs multi-objective optimization  \cite{dong2018dpp,scheidegger2020constrained}.
Instead of finding a set of solutions, some studies attempt to find the best tradeoff between efficiency and effectiveness. Two schemes are commonly used in this regard and mainly for the RL-based and differentiable algorithms:\textbf{multiplication} and \textbf{linear combination}. For the \textit{multiplication} scheme, the effectiveness objective (e.g., accuracy) and efficiency objective (e.g., latency) are multiplied to construct the loss function \cite{fbnet2019wu} or reward function \cite{howard2019searching,tan2019mnasnet,mnasfpn2020chen,li2021searching}:
\begin{equation}\label{multiplication_loss}
    \text{loss function: }\mathbfcal{L}(a)=\mathbfcal{L}_{CE}(w|a)\times \text{log}(C(a))^\beta,
\end{equation}
\begin{equation}
    \text{reward function: }\mathbfcal{R}(a)=ACC(a)\times (C(a)/C_0)^\beta.
\end{equation}
The $\mathbfcal{L}_{CE}(w|a)$ is the cross-entropy loss of model $a$ with parameter $w$. The exponent coefficient ($\beta>0$ for the loss function; $\beta<0$ for the reward function) modulates the trade-off between effectiveness and efficiency. Note that although there is a target constraint term $C_0$ in the reward function, this is still a soft constraint since this just pushes the cost to be lower than the target but without any guarantee. Alternatively, it is possible to keep running the search process until the hardware constraint is satisfied \cite{howard2019searching}. For the \textit{linear combination} scheme, the two competing objectives are linearly combined to construct the loss function \cite{singlepath2019,fang2020,wan2020fbnetv2,hrnas2021ding,proxyless} or reward function \cite{xiong2021mobiledets,DBLP:conf/cvpr/BenderLCCCKL20,chu2021discovering}:
\begin{equation}\label{linear_loss}
    \text{loss function: }\mathbfcal{L}(a)=\mathbfcal{L}_{CE}(w|a)+\beta\:\text{log}(C(a)),
\end{equation}
\begin{equation}
    \text{reward function: }\mathbfcal{R}(a)=ACC(a)+\beta|C(a)/C_0-1|.
\end{equation}
Similar to the multiplication scheme, the $\beta$ coefficient balances the effectiveness and efficiency. The log($\cdot$) function in the  hardware-related term in the loss function is used to scale the cost and are omitted in some studies \cite{wan2020fbnetv2,proxyless,hrnas2021ding}. TuNAS \cite{DBLP:conf/cvpr/BenderLCCCKL20} empirically demonstrates that search results are robust to the exact value of $\beta$. It shows that the same $\beta$ value works great for different search spaces and hardware constraints. This $\beta$ value-invariance alleviates the tedious tuning of $\beta$ for new scenarios (e.g., new devices). 

For non-differentiable search algorithms, the hardware constraint can be smoothly incorporated; by contrast, differentiable search algorithms require the cost term to be differentiable, which is intrinsically not. Therefore, specific cost estimation strategies are expected to resolve this contradiction, which we will introduce in Section \ref{predictor}. 


\subsection{Performance Estimation Strategy}\label{performance}
The search process is managed by search strategies, which we have discussed in Section \ref{searchstrategy}, and guided by the performance of candidate models, which we will discuss in this section. The performance in the context of hardware-efficient models has two aspects: task-related performance (e.g., accuracy) and hardware-related performance (e.g., latency).
The most natural way to obtain a model's performance is to fully train the model and evaluate it on validation data for task-related performance and deploy it on the target hardware for hardware-related performance (for direct metrics, e.g., latency) \cite{DBLP:conf/dac/OdemaRDF21,lu2019nsga}. However, it demands massive computation and time to fully train each candidate model from scratch and deploy them on target hardware. Therefore, efforts are made toward alleviating the performance estimation process. We categorize and discuss these efforts into two routes: one is to speed up the training process (section \ref{training}), and the other is to directly predict the performance without training (section \ref{predictor}). Since the hardware cost does not require training, studies of estimating hardware-related performance are reviewed in section \ref{predictor} as well.

\subsubsection{Training Strategy}\label{training}
The hardware-aware NAS does not distinguish from conventional NAS regarding the training strategies of candidate models so the training strategy summarized in this section can also be applied to conventional NAS and vice versa. The simplest training strategy is to train each candidate model from scratch and use the sample-eval-update loop \cite{DBLP:conf/iclr/ZophL17} to achieve an effective search agent. This strategy is not only widely adopted by RL-based search \cite{tan2019mnasnet,mnasfpn2020chen,howard2019searching,zhou2018resource,zoph2018learning}, but also by BayesOpt \cite{DBLP:conf/dac/OdemaRDF21,DBLP:journals/sensors/YangZLLWWZ21,parsa2020bayesian}, evolutionary search \cite{lu2019nsga,DBLP:conf/iclr/ElskenMH19,scheidegger2020constrained}, and other algorithms \cite{li2021searching,efficientnet}. To alleviate the drawback of the straightforward training approach, some studies \cite{mnasfpn2020chen,tan2019mnasnet,resource2019} follow Zoph et al. \cite{zoph2018learning} to first train and search on {\it proxy} tasks, such as smaller datasets or fewer epochs, then transfer to the large-scale target task. The proxy task is a reduced version of the target task. Xiong et al. \cite{resource2019} also propose {\it lazy evaluation} to further allow fewer evaluation. However, this is specific to its search strategy. An inevitable shortcoming of using a proxy task is that the model optimized on proxy tasks is not guaranteed to be favourable on the target task. 

Another way of speeding up the training process is {\it parameter sharing} \cite{DBLP:conf/icml/PhamGZLD18}. The search space is represented as a single super Directed Acyclic Graph (DAG), dubbed a {\it supernet}, whose nodes are computation operators and edges illustrate the flow of information. Each candidate model is a subgraph/subnet of the supernet and candidate models can share their parameters if they share common operators. When the supernet is well trained (e.g., trained for enough epochs), a candidate model can directly inherit parameters from the supernet without any separate training. A more widely used supernet is to have each node represent a latent representation (e.g., feature maps of a CONV operator) and each directed edge represents some operation (e.g., a CONV operator). There are multiple edges (i.e., different operators) between every pair of nodes, and the search process is to determine which edge/edges should be retained between each pair of nodes. The common retained edges share parameters between different candidate models. Different studies differ in training the supernet. Some work \cite{hsu2018monas,DBLP:conf/cvpr/BenderLCCCKL20,DBLP:conf/date/LuoLHL21} first pre-trains the whole supernet, and then only trains a part (e.g. a subnet) of the supernet, which is selected by search algorithms. Specifically, TuNAS \cite{DBLP:conf/cvpr/BenderLCCCKL20} disables the RL controller at the first 25\% of the search and only trains the parameters of the supernet. It randomly selects CNN filters and candidate operators to train with some probability $p$, which is linearly decreased from 1 to 0 over the first 25\% search. During the parameter sharing, TuNAS further shares the ``submodule'' parameters. For example, MBConv operators with different DWSConv kernel sizes are in different paths/edges, but they can share the PWConv weights in MBConv regardless of which DWConv kernel is selected. In addition, TuNAS also applies channel masking for parameter sharing; it creates a convolution with the largest possible number of channels while simulating smaller channel numbers by choosing the first $N$ channels while zeroing out the remaining ones. These techniques are well recognized and followed by many researchers \cite{chu2021discovering,xiong2021mobiledets}.

Differentiable NAS naturally uses a parameter-sharing training strategy. It jointly trains the supernet's parameters and the importance weights of each path/edge \cite{fbnet2019wu}. Furthermore, some researchers \cite{fang2020,hrnas2021ding} propose to progressively discard the paths/edges with low importance during training to further accelerate the training process. This progressive shrinking strategy speeds up the training process significantly. 
A defect of the differentiable parameter sharing strategy is that the supernet consumes high GPU memory, which would grow linearly with regard to the number of candidate models. To fill this gap, the single-path training strategy is proposed to reduce the memory cost to the same level as training a single candidate model. The conventional parameter sharing strategy has different operators on different paths, even though these operators are of the same type but with different hyperparameters (e.g., different kernel sizes). In light of this observation, Single-Path NAS \cite{singlepath2019} designs a superkernel $\textbf{w}_k$ for the DWConv inside a MBConv to choose between a $3\times3$ or a $5\times5$ kernel:
\begin{equation}\label{eq:singlepath}
    \textbf{w}_k = \textbf{w}_{3\times3}+\mathbbm{1}(||\textbf{w}_{5\times5\setminus3\times3}||^2>t_k)\cdot\textbf{w}_{5\times5\setminus3\times3}.
\end{equation}
It views the $3\times3$ kernel as the {\it inner} core of the $5\times5$ kernel, while zeroing out the weights of the {\it outer} shell. The $5\times5$ kernel can be viewed as the summation of this \textit{inner} core $\textbf{w}_{3\times3}$ and the \textit{outer} shell $\textbf{w}_{5\times5\setminus3\times3}$: $\textbf{w}_{5\times5}=\textbf{w}_{3\times3}+\textbf{w}_{5\times5\setminus3\times3}$. The  choice of the kernel size can be done using the indicator function $\mathbbm{1}(\cdot)\in\{0,1\}$ in Equation (\ref{eq:singlepath}), which is relaxed to be a sigmoid function $\sigma(\cdot)$ to compute gradients. The $t_k$ is a learned threshold. Likewise, the NAS decision of the expansion ratio $e\in\{3,6\}$ of MBConv can be represented as:
\begin{equation}\label{eq:superER}
\small
    \textbf{w}_{k,e}=\mathbbm{1}(||\textbf{w}_{k,3}||^2>t_{e=3})\cdot(\textbf{w}_{k,3}+\mathbbm{1}(||\textbf{w}_{k,6\setminus3}||^2>t_{e=6})\cdot\textbf{w}_{k,6\setminus3}),
\end{equation}
where $\textbf{w}_{k,3}$ is the channels with expansion ratio $e=3$, and can be viewed as the first half of the channels of the MBConv with expansion ratio $e=6$, while zeroing out the second half of the channels $\textbf{w}_{k,6\setminus3}$. The first indicator function in the Equation (\ref{eq:superER}) is used to decide whether to skip the MBConv layer. This NAS problem can then be formulated as a common {\it single-level} optimization problem:
\begin{equation}
    \min_\textbf{w}\mathbfcal{L}(\textbf{w}|t_k,t_e),
\end{equation}
as opposed to the {\it bi-level} optimization in Equation (\ref{eq:bilevel}). This optimization is solved in a differentiable way. The memory consumption of the single-path NAS is of the same level as the largest candidate model. Nevertheless, this single-path strategy only validates searching for different hyperparameters of the same type of operators, especially for convolution-based operators. 
ProxylessNAS \cite{proxyless}, on the other hand, still relies on a multi-path structure but only activates two paths at each training iteration, which are sampled with the highest learnable probabilities, and mask out all the other paths. FBNetV2 \cite{wan2020fbnetv2} combines the single-path and multi-path strategies. It also considers the smaller number of channels as part of the larger volume of channels via vector masks, each of which has ones in the first entries and zeros in the remaining entries; different vector masks are selected with Gumbel softmax weights; in this way, only the largest set of filters needs to be trained. For the resolution search, the authors propose to subsample smaller feature maps from the largest feature map, perform convolution, and enlarge with inserted zeros to the largest size. This design resolves the resolution mismatch problem during single-path training. Although FBNetV2 is still a multi-path structure, it reduces the network paths with the single-path strategy thus accelerating the training and reducing the memory cost.

Different from the joint optimization of gradient-based methods, evolutionary search or RL-based search decouples the training and search process into two sequential steps. Thus, it is unnecessary to use indicator functions or learnable weights in the training. The single-path training strategy becomes simpler. For example, SPOS \cite{DBLP:conf/eccv/GuoZMHLWS20} proposes to {\it uniformly sample} a single path from a supernet and makes it well trained. As a result, all candidate models are trained evenly and simultaneously. This strategy alleviates the co-adaptation problem of shared weights \cite{DBLP:conf/icml/BenderKZVL18}. After training, subnets can be directly sampled from the supernet for search evaluation. Many papers \cite{yan2021light,tang2020searching,DBLP:conf/iccv/Chu0X21,chu2020moga} then follow this simple yet effective strategy, and design novel techniques to improve fair sampling and training of candidate models \cite{tang2020searching,DBLP:conf/iccv/Chu0X21,chu2020moga}. A comprehensive study is FairNAS \cite{DBLP:conf/iccv/Chu0X21,chu2020moga} that ensures the parameters of every candidate operator be updated the same amount of times at any training iteration. Specifically, similar to SPOS, it randomly samples a candidate operator at each layer to form a subnet but differently, the chosen operators are not put back at the next sampling step. The sampling process continues until all operators are sampled. The sampled models are trained individually with back-propagation, but their gradients are accumulated to update the supernet’s parameters. This strategy can alleviate the {\it ordering issue} \cite{DBLP:conf/iccv/Chu0X21}, where candidate models are trained with an inherent training order in SPOS. Though SPOS and FairNAS accelerate the training process and reduce the memory cost, it is conventionally required to retrain the searched model before deployment because the inherited parameters from the supernet are not specialized for a specific model. Several studies \cite{DBLP:conf/eccv/YuJLBKTHSPL20,xu2021vipnas,cai2020once} attempt to form a {\it single-training} strategy that does not require any post training after searching. BigNAS \cite{DBLP:conf/eccv/YuJLBKTHSPL20,xu2021vipnas} employs a bunch of training techniques (e.g., sandwich rule, in place distillation, and batch norm calibration) \cite{DBLP:conf/iccv/YuH19,DBLP:conf/eccv/YuJLBKTHSPL20} to achieve a high quality supernet so the inherited parameters can work well for any subnets. OFA \cite{cai2020once} is a more promising strategy that prevents interference between subnets and thus a derived model can be directly deployed. It trains candidate models from the largest size (i.e., largest kernel size, depth, and width) to the smallest size (i.e., smallest kernel size, depth, and width) progressively. When training smaller subnets, the authors keep the last layers or channels untouched and finetune the early ones from shared parameters. As for elastic kernel size, the authors train kernel transformation matrices to transform the center of the larger kernel into the smaller kernel. The OFA strategy ensures each candidate model has a specifically trained or finetuned part so that mitigates their interference. 

\subsubsection{Performance Predictor}\label{predictor}
Another strategy to speed up the training process is to estimate the accuracy of candidate models using an \textbf{accuracy predictor} \cite{cai2020once,fbnetv32021dai,dong2018dpp,DBLP:conf/cvpr/DaiZWYSWDHWJVUJ19,DBLP:conf/iccv/MoonsNSMMLB21}. Although it requires substantial [model, accuracy] pairs thus computation and time to construct an accuracy predictor, it is a one-time cost as the predictor can be re-used for multiple hardware constraints. In addition, the accuracy predictor can also be easily finetuned for new datasets. The input into the accuracy predictor is the representation of candidate models, which is often one-hot encoding of candidate operators and hyperparameters \cite{cai2020once,dong2018dpp} or continuous values of hyperparameters (e.g., for channel counts) \cite{fbnetv32021dai,DBLP:conf/cvpr/DaiZWYSWDHWJVUJ19}. A different study is DONNA \cite{DBLP:conf/iccv/MoonsNSMMLB21}, which uses block-quality metrics derived from blockwise knowledge distillation as the input into the accuracy predictor. For the predictor architecture, some studies use neural networks (e.g., MLP \cite{fbnetv32021dai,cai2020once} or RNN \cite{dong2018dpp}) while others use conventional learning models (e.g., linear regressor \cite{DBLP:conf/iccv/MoonsNSMMLB21} or Gaussian Process regressor \cite{DBLP:conf/cvpr/DaiZWYSWDHWJVUJ19}). ChamNet \cite{DBLP:conf/cvpr/DaiZWYSWDHWJVUJ19} additionally adopts Bayesian optimization for model sampling to achieve better sampling efficiency and reliable prediction. 

In addition to an accuracy predictor, \textbf{predictors for hardware cost} are also widely studied to reduce the huge communication expense between the target device and the model training machine. Furthermore, the hardware cost predictor is demonstrated high fidelity across platforms ($r^2\geq0.99$) \cite{xiong2021mobiledets}. Therefore, the cost predictor is popular and necessary. A broadly investigated approach is the latency \textit{Lookup Table (LUT)} \cite{DBLP:conf/cvpr/BenderLCCCKL20,mnasfpn2020chen,fbnet2019wu,fang2020,singlepath2019,DBLP:conf/date/LuoLHL21,chu2020moga,DBLP:conf/cvpr/DaiZWYSWDHWJVUJ19,cai2020once}, which records the runtime of each operator in the search space. The basic assumption is that the runtime of each operator is independent of other operators \cite{fbnet2019wu,proxyless}, so that the latency of an entire model $a$ can be estimated as the sum of the latency of each individual operator $O_i$:
\begin{equation}
    LAT(a)=\sum_i LAT(O_i).
\end{equation}
Researchers profile the target hardware and record the runtime for each candidate operator to estimate the latency of candidate models. Some research also considers the connectivity of operators and the communication overheads in sequential layers \cite{fang2020,DBLP:conf/date/LuoLHL21,mnasfpn2020chen}. 
Another possible way is to construct a regressor to estimate the hardware cost based on critical features of a candidate model \cite{DBLP:conf/dac/OdemaRDF21,proxyless,DBLP:conf/cvpr/DaiZWYSWDHWJVUJ19,xiong2021mobiledets,chu2021discovering}. The performance regressor can be a linear regressor \cite{xiong2021mobiledets,chu2021discovering}, a Gaussian Process regressor \cite{DBLP:conf/cvpr/DaiZWYSWDHWJVUJ19}, or a neural network \cite{DBLP:conf/dac/OdemaRDF21,proxyless}. Different from directly measuring the hardware cost and training a black-box performance regressor, two papers examine specific devices and derive the runtime \cite{DBLP:journals/sensors/YangZLLWWZ21} or power consumption \cite{parsa2020bayesian} through theoretical analysis. Nevertheless, this approach requires the knowledge of specific devices and is inflexible to different devices. 

The hardware cost predictor is essential for differentiable search algorithms. Since each candidate operator is to be selected by a binary indicator in a differentiable supernet, the latency of it can be the weighted sum (i.e., with binary indicators \{0,1\}) of the latency of each candidate operator \cite{fbnet2019wu}:
\begin{equation}\label{latency_diff}
    LAT(a)=\sum_l\sum_i \mathbb{I}_{l,i}\cdot LAT(O_{l,i}), \;\mathbb{I}_{l,i}\in \{0,1\},
\end{equation}
where $O_{l,i}$ and $\mathbb{I}_{l,i}$ are the $i$th operator of the $l$th layer and its associated binary indicator, respectively. $LAT(O_{l,i})$ can be achieved through either LUT \cite{fbnet2019wu,singlepath2019,fang2020} or performance regressor \cite{proxyless}. However, the loss function (\ref{multiplication_loss}) and (\ref{linear_loss}) with indicator $\mathbb{I}_{l,i}$ is not directly differentiable. To sidestep this problem, the indicator is relaxed to be a continuous variable computed via the \textit{Gumbel Softmax} function \cite{DBLP:conf/iclr/JangGP17} with learnable parameters:
\begin{equation}\label{indicator}
    \mathbb{I}_{l,i}=\frac{exp[(\theta_{l,i}+g_{l,i})/\tau]}{\sum_i exp[(\theta_{l,i}+g_{l,i})/\tau]},
\end{equation}
where $\tau\in(0,1)$ is the temperature parameter that controls the search efficiency and efficacy. Larger $\tau$ makes the indicator distribution smoother; smaller $\tau$ approaches discrete categorical sampling. Despite the value of $\tau$, the loss function (\ref{multiplication_loss}) and (\ref{linear_loss}) with the latency calculation (\ref{latency_diff}) and the relaxed indicator (\ref{indicator}) are differentiable. For single-path training, the latency calculation (\ref{latency_diff}) can be simplified \cite{proxyless} as:
\begin{equation}
    LAT(a)=\sum_l \mathbb{I}_{l}\cdot LAT(O_{l}), 
\end{equation}
where $O_{l}$ is the selected operator at layer $l$ and $\mathbb{I}_{l}$ is the path probability of operator $O_l$.

The cost of preparing a hardware cost predictor is minimal since it does not require training models. Only one forward pass of a test model is sufficient to record its cost.

\begin{table*}[]
    \centering
    \input{table/search}
    \label{tab:search}
\end{table*}

\begin{table*}[]
\ContinuedFloat 
    \centering
    \input{table/search_cont}
    \label{tab:searchcont}
\end{table*}

%% file: table/search.tex
\fontsize{5.3pt}{8pt}\selectfont
\caption{Summary of Searching for Efficient Deep Learning Models. The ``hardware'' column indicates on which the achieved models are evaluated. The latency is reported per input; otherwise is specified in the relevant table cells (e.g., FPS). $\sim$ indicates the exact data is not reported in the paper and thus estimated from the reported figures. `-' indicates unavailable records.}
\begin{tabular}{@{}>{\centering}p{0.6cm}|c|c|c|c|c|c|c@{}}
\toprule
Search strategy                                & Reference                                                                   & Dataset                         & Hardware                       & Accuracy                                                               & Latency                                          & FLOPs                                        & \# of Parameters                   \\ \midrule
\multirow{4}{*}{\rotatebox[origin=c]{90}{\scriptsize{BayesOpt+Soft}}}        & \multirow{4}{*}{\cite{DBLP:journals/sensors/YangZLLWWZ21}} & \multirow{2}{*}{MNIST}          & Xilinx Zynq-7000 SoC ZC70      & acc: 99.49                                                             & 64.74ms                                          & \multirow{4}{*}{-}                           & \multirow{4}{*}{-}                 \\ \cmidrule(lr){4-6}
                                      &                                                                             &                                 & Xilinx Virtex-7 FPGA VC707     & acc: 99.46                                                             & 105.54ms                                         &                                              &                                    \\ \cmidrule(lr){3-6}
                                      &                                                                             & \multirow{2}{*}{CIFAR10}        & Xilinx Zynq-7000 SoC ZC70      & acc: 88.25                                                             & 136.15ms                                         &                                              &                                    \\
\bottomrule
\end{tabular}

%% file: table/search_cont.tex
\fontsize{5.3pt}{8pt}\selectfont
\caption{Summary of Searching for Efficient Deep Learning Models (cont.).}
\begin{tabular}{@{}>{\centering}p{0.6cm}|c|c|c|c|>{\centering}p{1.3cm}|c|c@{}}
\toprule
Search strategy                                & Reference                                                                   & Dataset                         & Hardware                       & Accuracy                                                               & Latency                                          & FLOPs                                        & \# of Parameters                   \\ \midrule

\multirow{11}{*}{\rotatebox[origin=c]{90}{\footnotesize{Other+Soft}}}           & \multirow{7}{*}{\cite{dong2018dpp}}                        & \multirow{5}{*}{CIFAR10}        &                                & error: 4.62                                                            & 9ms/82ms/149ms                                   & 63.5M                                        & 0.52M                              \\ \cmidrule(lr){5-8} 
                                      &                                                                             &                                 & Titan X/                  & error: 4.36                                                            & 13ms/62ms/912ms                                  & 1364M                                        & 11.39M                             \\ \cmidrule(lr){5-8} 
                                      &                                                                             &                                 & Maxwell 256/                   & error: 4.78                                                            & 6ms/75ms/210ms                                   & 137M                                         & 1.00M                              \\
\bottomrule
\end{tabular}

%% file: content/4.automated_compression.tex
\section{Automated Compression of Deep Learning Models}\label{section:compression}
Since deep learning models have proved remarkable performance in a wide spectrum of problems, it is a natural idea to compress these well-established models for memory saving and compute acceleration and thus hardware-efficiency \cite{DBLP:journals/pieee/DengLHSX20}. This idea is fervently supported by the fact that the hand-crafted deep learning models are often over-parameterized \cite{DBLP:journals/air/ChoudharyMGS20,DBLP:conf/nips/DenilSDRF13,LeCun1989nips}. The goal of deep learning compression is to modify well-trained models for efficient execution without significantly compromising accuracy. Although related works are quite divergent, they can be broadly summarized into four categories \cite{DBLP:journals/air/ChoudharyMGS20,DBLP:journals/pieee/DengLHSX20}: tensor decomposition, knowledge distillation, pruning, and quantization. In the past few years, the above compression techniques have achieved great success while they crucially rely on domain knowledge, hand-crafted designs, and tremendous efforts for tuning. Recently, there is a growing demand and trend for automating the compression process on all the above four compression categories. In this section, we aim to provide a comprehensive review of recent research studies on automated compression of neural architectures with regard to the four compression categories.

\subsection{Automated Tensor Decomposition}
As the computation in neural networks is based on tensor operations, it is intuitive to compress tensors to squeeze and accelerate a neural network. The basic operation of tensor decomposition is the mode-$i$ product of a tensor with a matrix. For a $d$th-order tensor ${\mathbfcal{X}}\in\mathbb{R}^{n_1\times n_2\times ... \times n_d}$ and a matrix $\textit{\textbf{B}}\in\mathbb{R}^{m\times n_i}(i\in\{1,2,...,d\})$, the mode-$i$ product between $\mathbfcal{X}$ and $\textit{\textbf{B}}$ is $\mathbfcal{R}=\mathbfcal{X}\times_i\textit{\textbf{B}}$, where $\mathbfcal{R}\in\mathbb{R}^{n_1\times ... \times n_{i-1}\times m\times n_{i+1} ... \times n_d}$ is also a $d$th-order tensor. Given this definition, a $d$th-tensor $\mathbfcal{A}\in\mathbb{R}^{n_1\times n_2\times ... \times n_d}$ can be decomposed into one core $d$th-order tensor $\mathbfcal{G}\in\mathbb{R}^{r_1\times r_2\times ... \times r_d}$ and $d$ factor matrices $\textbf{\textit{U}}^{(i)}\in\mathbb{R}^{n_i\times r_i}(i\in{1,2, ..., d})$:
\begin{equation}
    \mathbfcal{A}\approx\mathbfcal{G}\times_1\textbf{\textit{U}}^{(1)}\times_2\textbf{\textit{U}}^{(2)}\times_3 ... \times_d\textbf{\textit{U}}^{(d)}.
\label{decomp}    
\end{equation}
This decomposition is called \textit{Tucker decomposition} \cite{tucker1966some}, and the tuple $(r_1, r_2, ..., r_d)$ is called the \textit{Tucker rank}. By selecting proper low ranks (i.e., $r_i<n_i$), the original tensor can be represented by lightweight decomposed pieces. The parameter compression ratio is:
\begin{equation}
    P=\frac{\prod_{i=1}^{d}n_{i}}{\sum_{i=1}^{d}r_{i}n_{i}+\prod_{i=1}^{d}r_{i}}.
\end{equation}
Given that the parameters of a neural network are in the form of tensors (e.g., the filter in a convolutional layer is a 4-way tensor), tensor decomposition can be naturally applied to model the parameters in a more efficient way. The speed-up ratio of a convolution layer is:
\begin{gather}
\scalebox{1}{$
S=\frac{n_{1}n_{2}n_{3}n_{4}H'W'}{n_{1}r_{1}H'W+n_{2}r_{2}HW'+n_{3}r_{3}HW+n_{4}r_{4}H'W'+
    r_{1}r_{2}r_{3}r_{4}H'W'}
\nonumber$},
\end{gather}
where $H\times W$ is the input feature map size, $H'\times W'$ is the output feature map size, and $n_{1}\times n_{2}\times n_{3}\times n_{4}$ is the kernel size. 
Tucker decomposition is a widely applied approach to this aim by tensorization of neural network layers \cite{DBLP:journals/corr/KimPYCYS15,DBLP:conf/cvpr/KossaifiKLFA17}. Another commonly used tensorization approach is the Canonical Polyadic (CP) decomposition \cite{carroll1970analysis}, which is a special case of the Tucker decomposition \cite{DBLP:conf/bigcomp/AstridL17,DBLP:journals/corr/LebedevGROL14}. After  decomposition, it is usually required to fine-tune the tensorized network to recover the accuracy.

When tensorizing neural networks, the decomposition rank is the most important hyperparameter that needs to be carefully selected since it controls the compression-accuracy trade-off. A typical selection method is cross-validation, which is quite cumbersome for selecting a diverse range of ranks, so the common practice is to set the ranks of different layers to be the same. This simplification is coarse and sub-optimal. Therefore, automating the rank-selection process is crucial to determining optimal decomposition ranks. Kim et al. \cite{DBLP:journals/corr/KimPYCYS15} propose to employ global analytic solutions for variational Bayesian matrix factorization (VBMF) \cite{DBLP:journals/jmlr/NakajimaSBT13} for automatic rank selection of Tucker decomposition. Publicly available tools can easily implement the VBMF. They perform full model compression including fully connected and convolutional layers and show that the accuracy degradation after compression can be well recovered by fine-tuning. However, Gusak et al. \cite{DBLP:conf/iccvw/GusakKPMBCO19} claim that they find it difficult to restore the initial accuracy by fine-tuning with the global analytic VBMF ranks. Therefore, they use the global analytic solution of Empirical VBMF (EVBMF) to automatically select ranks. Instead of directly using the ranks achieved from EVBMF, the authors design a \textit{weakened rank} for decomposition, which is larger than the \textit{extreme rank} (i.e., obtained by EVBMF) and thus not optimal with a certain amount of redundancy after decomposition. The reason for this design is that this work proposes an iterative compression and fine-tuning strategy that gradually compresses the original model and restores the initial accuracy. 
Different from tensor decomposition on a well-trained network followed by fine-tuning in \cite{DBLP:journals/corr/KimPYCYS15,DBLP:conf/iccvw/GusakKPMBCO19}, Hawkins et al. \cite{hawkins2022towards,hawkins2021bayesian} operate automatic rank determination and tensor decomposition along with the model training process. They propose a prior distribution mask, which can be optimized  during training, over the decomposition core tensor to control the rank. If an element of the resultant posterior mask is smaller than a threshold, it will be regarded as zero and the rank will be automatically selected. The authors use a Bayesian inference method to train this low-rank tensorized model and prove its efficacy for various decomposition schemes. MARS \cite{kodryan2020mars} sets a factorized Bernoulli prior and optimizes the model via relaxed MAP (maximum a posteriori) estimation to achieve a binary mask to control the rank so the manually set threshold used in \cite{hawkins2022towards,hawkins2021bayesian} is avoided. It shows slightly worse in compression but better accuracy than \cite{hawkins2021bayesian}.
Inspired by the success of reinforcement learning in making decisions, Javaheripi et al. \cite{javaheripi2021autorank,samragh2019autorank} propose a state-action-reward system to automatically select the optimal rank for each layer. The hardware cost and accuracy are both considered in the reward and the action of decomposing a layer with the highest rewarded rank is selected. Bayesian optimization can also utilized to achieve the optimal ranks \cite{DBLP:conf/iccv/MaTBSCB19}.

TABLE \ref{tab:td} summarizes the main results of automated tensor decomposition studies. Note that the latency is achieved with the hardware listed in the table, which is different from the hardware for training. 
\begin{table*}[ht]
    \centering
    \input{table/TD}
    \label{tab:td}
\end{table*}

\subsection{Automated Knowledge Distillation}
Knowledge distillation (KD) is extended from knowledge transfer (KT) \cite{DBLP:conf/kdd/BucilaCN06} by Ba and Caruana \cite{DBLP:conf/nips/BaC14} to compress a cumbersome network (teacher) into a smaller and simpler network (student). This is done by making the student model mimic the function learned by the teacher model in order to achieve a competitive accuracy. It is later formally popularized by Hinton et al. \cite{hinton2015distilling} as a student-teacher paradigm, where the knowledge is transferred from the teacher to the student by minimizing the difference between the logits (features before the final softmax) of the teacher and student. In many situations, the performance of the teacher is almost perfect with a very high classification probability for the correct class and flat probabilities for the other classes. Therefore, the teacher is not able to provide much more information than the ground truth labels. Hinton et al. \cite{hinton2015distilling} introduce the concept of \textit{softmax temperature} to transfer knowledge, which can better deliver the information of which classes the teacher find similar to the correct class. Formally, given the logits of the teacher model, the classification probability $p_i$ of the class $i$ is:
\begin{equation}\label{eq:temp}
    p_i=\frac{exp(\frac{z_i}{\tau})}{\sum_j exp(\frac{z_i}{\tau})},
\end{equation}
where $\tau$ is the \textit{temperature} parameter. It controls how \textit{soft} the labels from the teacher are. The soft labels together with the ground truth labels are used to supervise a compact student model.

Vanilla knowledge distillation mostly focuses on transferring knowledge to a student model with a fixed small architecture, which is manually designed in advance. However, different teachers and tasks favour different student architectures, and hand-crafted architectures are prone to be sub-optimal. Considering these limitations, there is a growing trend to automate the architecture design of a student model \cite{DBLP:conf/cvpr/LiuJTVZGW20,guo2021rosearch,DBLP:journals/algorithms/LeiYJC21,eyono2021autokd,zhang2022autodistill,chen2021scene,DBLP:conf/icpr/ZhangZYGX20}. The ground-truth labels are combined with the distillation labels to guide the automatic design process.
AKDNet \cite{DBLP:conf/cvpr/LiuJTVZGW20} proposes to search optimal student architectures for distilling a given teacher by RL-based NAS. It adopts the efficient search space of \cite{tan2019mnasnet} and designs a KD-guided reward with a teacher network. KDAS-ReID \cite{DBLP:journals/algorithms/LeiYJC21}, AdaRec \cite{chen2021scene}, and NAS-KD \cite{DBLP:conf/icpr/ZhangZYGX20} leverage the differential NAS \cite{liu2018darts} to determine the optimal student structures. In addition, Bayesian optimization \cite{zhang2022autodistill,eyono2021autokd} and evolutionary search \cite{guo2021rosearch} are also popular search strategies in the context of student architecture search. 

Although the soft labels from the final output of a teacher have been demonstrated effective for transferring knowledge, some works argue that it is also helpful to mimic the teacher from the intermediate layers. AdaRec \cite{chen2021scene} and NAS-KD \cite{DBLP:conf/icpr/ZhangZYGX20} try to minimize the difference between the intermediate features of the student and the teacher. Similarly, Li et al. \cite{DBLP:conf/cvpr/LiLDLZH20} and MFAGAN \cite{cheng2021mfagan} compress the student generator in Generative Adversarial Networks (GANs) via intermediate feature distillation and \textit{once-for-all} NAS \cite{cai2020once}. PPCD-GAN \cite{vo2022ppcd} also compresses the generator but with a teacher-guided learnable mask to automatically reduce the number of channels.  

Unlike searching for how to reduce, Kang et al. \cite{DBLP:conf/aaai/KangMH20} and Mitsuno et al. \cite{DBLP:conf/icpr/MitsunoNK20} propose to search on how to increase. They set an extremely small student backbone at the start point, and augment operations \cite{DBLP:conf/aaai/KangMH20} or additional layer channels \cite{DBLP:conf/icpr/MitsunoNK20} to the backbone during the search procedure. A large pre-trained teacher \cite{DBLP:conf/icpr/MitsunoNK20} or an ensemble teacher \cite{DBLP:conf/aaai/KangMH20} is used to guide the search process. The motivation of these works is to alleviate the search burden with a start point and to maximize the distilled knowledge by optimally reducing the capacity gap between teacher and student.

A different direction is NAS-BERT \cite{DBLP:conf/kdd/Xu0LS0QL21}, where the block-wise KD \cite{DBLP:conf/cvpr/LiPYWLLC20} is applied to train a supernet composing a bunch of candidate compact subnets. Then, the meta information (e.g., parameter, latency) of the candidate networks is summarized in a lookup table. Given certain hardware constraints, all candidate networks that satisfy the constraints are evaluated and the best performing one is selected as the final compressed model. The benefit of this strategy is that no retraining and researching are required for new hardware constraints.

Not only the student architecture is searchable, but also related hyperparameters (e.g., the temperature $\tau$ in Equation (\ref{eq:temp})) \cite{eyono2021autokd} can be included in the search process. AutoKD \cite{eyono2021autokd} uses the BayesOpt to simultaneously explore the optimal student structure and KD hyperparameters, temperature $\tau$ and loss weight $\alpha$, during the KD process. 

We outline the major results of automated knowledge distillation papers in TABLE \ref{tab:kd}. 

\begin{table*}[!ht]
    \centering
    \input{table/KD}
    \label{tab:kd}
\end{table*}

\subsection{Automated Pruning}\label{ssec: AP}
Neural network pruning, a.k.a. sparsification, attempts to reduce memory storage and computation cost by removing unimportant weights or neurons from the base network. According to the granularity, pruning can be roughly categorized into structured pruning and unstructured pruning. Unstructured pruning enforces weights or layers to be sparse by removing individual connections or neurons. Structured pruning, on the other hand, targets discarding the entire channels or layers. Although unstructured pruning can theoretically achieve a better accuracy-compression tradeoff, it is less compatible with existing deep learning platforms and hardware than structured pruning. Thus, structured pruning attracts more research focus. 

A typical pruning procedure is to first identify and cut away unimportant weighs or layers from an existing over-parameterized network and then fine-tune the pruned network to restore the accuracy. The first step is the key and largely relies on expert knowledge and hand-crafted heuristics, such as manually setting the criterion of {\it parameter importance} \cite{DBLP:conf/iclr/ParkLMS20} and the {\it pruning rate} \cite{DBLP:conf/iclr/ZhangYZWF18}. However, heuristic settings are typically based on trial and error and are sub-optimal so researchers actively seek automated solutions to alleviate human intervention during pruning. These efforts can roughly be categorized into three directions.

\textbf{1) Reducing the number of hyperparameters (i.e., per-layer pruning rates) needed to be tuned} \cite{zeng2021network,li2020ss,zheng2021information}.  
Zeng et al. \cite{zeng2021network} compute the importance of model parameters using the Taylor expansion of the loss function and prune unimportant ones. They introduce an auxiliary hyperparameter that controls the model shrinking proportion at each pruning iteration and show that this hyperparameter is insensitive to the final performance. In this way, they avoid carefully tuning the overall pruning rate. 
Li et al. \cite{li2020ss} use the regularization approach, \textit{Alternating Direction Method Multipliers} (ADMM), as the core pruning algorithm but substitute its hard constraint with the soft constraint-based formulation and solve the optimization problem with the Primal-Proximal solution. Their approach naturally does not require predefining the per-layer pruning rates and thus reduces the number of hyperparameters. Zheng et al. \cite{zheng2021information} introduce the normalized Hilbert-Schmidt Independence Criterion (nHSIC) from the information theory to measure the per-layer importance and derive the compression problem with constraints into a linear programming problem, which is convex and easily solved. In such a manner, only two hyperparameters are required and demonstrated robust to different values.

\textbf{2) Searching for optimal pruning rates or magnitude thresholds for pruning} \cite{He_2018_ECCV,DBLP:conf/aaai/0007MXW0Y20,DBLP:conf/ics/YangCZHYS21,DBLP:conf/mm/LiF0BZ21,DBLP:conf/bmvc/TungMM17,DBLP:conf/eccv/ChenTVM18,mu2021high,li2021abcp,DBLP:conf/dicta/GuerraD21,netadapt,yang2021netadaptv2}. 
Note that the magnitude thresholds of weights can be naturally derived from pruning rates.
AMC \cite{He_2018_ECCV} and Auto-Prune \cite{DBLP:conf/ics/YangCZHYS21} leverage deep reinforcement learning (DRL) to automatically determine the pruning rate of each layer and empirically prune the weights with the least magnitude. Arguing that DRL has an inherited incompatibility with the pruning problem, Liu et al. \cite{DBLP:conf/aaai/0007MXW0Y20} suggest using the heuristic search technique, \textit{simulated annealing} (SA), to search for the per-layer pruning rates and then use the ADMM to dynamically regularize network weights for structural pruning. They hypothesize that the layers with more weights can have higher compression rates. The authors also apply the ADMM to searching for the magnitude thresholds for the second-phase unstructured pruning after the first-phase structural pruning. 
Tung et al. \cite{DBLP:conf/bmvc/TungMM17} propose to employ a Bayesian optimization framework to search the pruning hyperparameters including the magnitude threshold.  Some studies \cite{DBLP:conf/eccv/ChenTVM18,mu2021high} follow this work to make it constraint-aware \cite{DBLP:conf/eccv/ChenTVM18} and applicable to pruning deeper networks \cite{mu2021high}. Other than pruning full-precision neural networks, Guerra and Drummond \cite{DBLP:conf/dicta/GuerraD21} aim at automatically pruning quantized neural networks by a customized, rule-based pruning strategy with Bayesian optimization of layer-wise pruning ratios.

\textbf{3) Learning the weight/channel importance}
\cite{guan2020dais,xiao2019autoprune,DBLP:conf/iccv/LiuLSHYZ17,DBLP:conf/eccv/LiGZGT20,DBLP:conf/dac/QuLWZ21}. Different from relying on heuristic weight/channel importance measurement like the weight magnitude, this direction struggles to learn the importance measurement through optimizing the final objective.  
Liu et al. \cite{DBLP:conf/iccv/LiuLSHYZ17} propose a slimming scheme that regularizes the scaling factors in batch normalization (BN) layers as a channel selection indicator to identify unimportant channels (or neurons). Then, they apply a threshold to the trained scaling factors for channel pruning. AutoPrune \cite{xiao2019autoprune} and DAIS \cite{guan2020dais} are following works that rely on auxiliary learnable channel selection indicator for pruning. AutoPrune argues that decoupling the channel selection indicators and network parameters will help to stabilize the weight learning and make the pruning insensitive to hyperparameters. DHP \cite{DBLP:conf/eccv/LiGZGT20} introduces an additional hypernetwork that can generate weights for the backbone network (i.e., the network to be pruned). The input of the hypernetwork is the latent vectors, which are attached to each layer of the backbone network. Binary channel masks will then be achieved by setting those latent vector elements, which are smaller than a threshold, to be zero otherwise to be one. In contrast to learning channel importance, ASBP \cite{DBLP:conf/dac/QuLWZ21} targets to automatically decide the binary importance of each bit of weight. This bit-level pruning has the finest granularity but is impractical to achieve a real reduction of resource consumption due to poor hardware support. To solve this problem, the authors target a specific sort of hardware, RRAM, which assembles multiple low-precision cells together as a crossbar to represent a high-precision data value. They then prune at the granularity of crossbar size, that is all cells in a crossbar share the same pruning strategy. The DL algorithm, deep deterministic policy gradient (DDPG), is used to search for the optimal set of bit-pruning strategies. 

\subsubsection{Unifying knowledge distillation and pruning} Since pruning and knowledge distillation both require a pre-trained model (i.e., the base model in pruning; teacher in KD) to guide the compression process, researchers naturally explore ways to unify these two techniques. One route is the two-step unifying scheme \cite{DBLP:conf/nips/Dong019,DBLP:conf/ai4i/XueX21}. TAS \cite{DBLP:conf/nips/Dong019} is an early and representative work. It first searches for the shrunken width and depth of the pruned network with the help of learnable probability from the base model. The searched architecture is then trained from scratch as the student of a simple KD approach with the base model serving as the teacher. The other route is to unify pruning and KD in a single step \cite{DBLP:conf/cvpr/YaoPXZLZ21,DBLP:conf/ecai/GuT20}. Gu and Tresp \cite{DBLP:conf/ecai/GuT20} introduce learnable channel gates into the base model and train the network weights and channel gates with a distillation-aware loss function. Thus, this scheme prunes the base model and distils its knowledge simultaneously and automatically. Yao et al. additionally search for both teacher and student to pursue the optimal distillation pair.

\begin{table*}[]
    \centering
    \input{table/AP}

    \label{tab:ap}
\end{table*}
\begin{table*}[]
\ContinuedFloat 
    \centering
    \input{table/AP_cont}

    \label{tab:apcont}
\end{table*}

\subsection{Automated Quantization}
Unlike the above three compression techniques, which struggle to optimize network architectures, quantization appeals to reduce the representation precision of network weights and intermediate activation tensors. Neural networks are generally trained using the 32-bit floating-point precision; if we were to perform network inference in the 32-bit floating-point as well, MAC operations, data transfer, and data saving would have to be done all in 32-bit floating-point. Hence, using lower bit precision would substantially reduce the hardware overhead, including communication bandwidth, computation and memory usage \cite{DBLP:conf/isscc/Horowitz14}. For example, when moving from 32 to 8 bits, the memory cost and computation would decrease by a factor of 4 and 16 respectively. In addition, fixed-point computation is more efficient than its floating-point counterpart \cite{DBLP:conf/isscc/Horowitz14}. The most commonly explored quantization scheme, \textit{uniform quantization}, converts a floating point tensor $\boldsymbol{x}=\{x_1, ... , x_N\}$ with range $(x_{min}, x_{max})$ into its integer coding $\boldsymbol{x}_q$ with range $[n, p]=[0, 2^b-1]$ via the following definition \cite{bai2021batchquant}:
\begin{equation}
    \boldsymbol{x}_q = \lfloor\text{clamp}(\frac{\boldsymbol{x}}{\Delta}+z; n, p)\rceil,
    \label{quantizer}
\end{equation}
\begin{equation}
      \Delta=\frac{x_{max}-x_{min}}{p-n}, \: z=\text{clamp}(-\lfloor\frac{x_{min}}{\Delta}\rceil+z; n, p),
\end{equation}
\begin{equation}
\boldsymbol{\hat{x}}=(\boldsymbol{x}_q-z)\Delta,
\end{equation}
where $\Delta$ is the \textit{scale factor} that specifies the step size of the quantizer; $z$ is the \textit{zero-point} that represents the real value zero without any precision loss; $b$ is the quantization \textit{bitwidth}; $\lfloor\cdot\rceil$ is the round-to-nearest operation. The $clamp(\cdot)$ function is to truncate all values to fall between $n$ and $p$. 
Through this procedure, two sorts of errors are induced: a \textit{clipping error} induced by the \textit{clamp} function and a \textit{rounding error} induced by the \textit{round-to-nearest} operation $\lfloor\cdot\rceil$. The quantized tensor $\boldsymbol{\hat{x}}$ is then used for efficient but low precision computation. 
Since there are no theoretical correlations between the model accuracy and $b$, simple try-and-error is only feasible in the scenario where all weights and activation tensors have the same precision. However, different layers have different redundancy and behaviours on both the hardware and the task. Therefore, automatically determining the bitwidth of weights and activations for each layer is stunningly attractive, and this mixed-precision feature has recently been supported by hardware manufacturers \cite{apple,nvidia}.

One direction is to model the layer-wise bit-precision assignment as a reinforcement learning problem, where the action is to determine the bitwidth values \cite{wang2019haq,DBLP:journals/ijcv/WangLLLH20}. HAQ \cite{wang2019haq,DBLP:journals/ijcv/WangLLLH20} is such a representative work. It utilizes the DDPG agent to explore a continuous action space of [0, 1], from which selected actions (i.e., real numbers) can be rounded into discrete bitwidth values for the weights and activations of each layer. The use of a continuous action space is based on the consideration that a discrete action space is not able to represent the relative order of bitwidth. The quantized model is then retrained one more epoch to recover the accuracy. Since the authors induce the hardware constraints by limiting the search space, they only consider minimizing the accuracy drop after quantization in the reward function. Another direction is to develop a differentiable framework that searches for the quantization bitwidth in the same way as searching for the network architecture \cite{DBLP:conf/uai/LaceyTA18,DBLP:conf/eccv/YuHLSCF20,DBLP:conf/icassp/XuHYLM21,sun2021effective,DBLP:journals/lgrs/WeiCLX21,wu2018mixed}.
Given a network to be quantized, a supernet is first constructed by inserting multiple parallel quantization operations into every pair of adjacent layers of the network. Each inserted operation constitutes an edge from the $i$-th layer to the ($i+1$)-th layer and is associated with a learnable parameter. Multiple parallel edges/operations on the same level are of different quantization bitwidths and summarized by assembling all edges with the weights derived from the learnable parameters. Then, the network parameters and the quantization weights are jointly trained with the combination of accuracy and model size as the target. Wu et al. \cite{wu2018mixed} and Xu et al. \cite{DBLP:conf/icassp/XuHYLM21} simply try to diminish the model size while Yu et al. \cite{DBLP:conf/eccv/YuHLSCF20} and Wei et al. \cite{DBLP:journals/lgrs/WeiCLX21} consider hardware constraints in the training target. SSPS \cite{sun2021effective} further borrows the idea from \cite{proxyless} to use a one-hot mask instead of continuous weights to select the bitwidth. In this way, only one path is active during training so the search cost is considerably reduced. The bitwidth search space is usually below eight since a previous study has shown that simple 8-bit post-training quantization is able to achieve marginal accuracy degradation \cite{nagel2021white}. We summarize the main studies of automated quantization in TABLE \ref{tab:aq}.

\begin{table*}[]
    \centering
    \input{table/AQ}
    \label{tab:aq}
\end{table*}

%% file: table/TD.tex

\fontsize{5.5pt}{8pt}\selectfont
\caption{Summary of Automated Tensor Decomposition Studies. The ``hardware'' column indicates on which the achieved models are evaluated. The model name (e.g., AlexNet) in the ``Accuracy'' column indicates the original model before tensor decomposition. Some studies only report relative performance changes which are denoted as $n\times$ in the table, indicating the original model is $n$ times its compressed counterpart. The latency is reported per input. `-' indicates unavailable records.}
\begin{tabular}{@{}c|c|c|c|c|c|c|c@{}}

\toprule
Method                                           & Reference                                             & Dataset                    & Hardware                 & Accuracy                                                                           & FLOPs                                 & Latency                     & \# of Parameters                   \\ \midrule
\multirow{25}{*}{\rotatebox[origin=c]{90}{\small{Automated Tensor Decomposition}}} & \multirow{2}{*}{\cite{DBLP:journals/corr/KimPYCYS15} }                 & \multirow{2}{*}{ImageNet} & Nvidia Titan X           & AlexNet/VGG-S/GoogleNet/VGG-16 (Top5 acc):                         & \multirow{2}{*}{272M/549M/760M/3139M} & 0.30ms/0.92ms/1.48ms/4.58ms & \multirow{2}{*}{11M/14M/4.7M/127M} \\ \cmidrule(lr){4-4} \cmidrule(lr){7-7}
                               &                                                       &                            & Samsung Galaxy S6        & 78.33/84.05/88.66/89.40                                       &                                       & 43ms/97ms/192ms/576ms       &                                    \\ \cmidrule(lr){2-8} 
                                                 & \multirow{3}{*}{\cite{DBLP:conf/iccvw/GusakKPMBCO19}}               & VOC2007                    & \multirow{5}{*}{-}       &  Faster R-CNN (mAP): 69.2/68.3/77.0/75.0                      & 10.49$\times$/13.95$\times$/1.57$\times$/1.49$\times$          & \multirow{5}{*}{-}          & \multirow{5}{*}{-}                 \\ \cmidrule(lr){3-3} \cmidrule(lr){5-6}
                                                 &                                                       & \multirow{2}{*}{COCO2014}  &                          &  Faster R-CNN FPN (mAP/mAP.50): 35.4/56.2                                         & 1.8$\times$          &                             &                                    \\ \cmidrule(lr){5-6}
                                                 &                                                       &                            &                          &  Faster R-CNN FPN (mAP/mAP.50): 36.2/57.1                                  &          1.7$\times$                            &                             &                                    \\ \cmidrule(lr){2-8} 
                                                 & \multirow{3}{*}{ARD-LU\cite{hawkins2022towards}  }                     & MNIST                      & \multirow{3}{*}{-}       &  MLP (Acc): 98.06/98.30/96.28/98.24                     & \multirow{3}{*}{-}                    & \multirow{3}{*}{-}          & 7k/101k/4k/6k                      \\ \cmidrule(lr){3-3} \cmidrule(lr){5-5} \cmidrule(lr){8-8} 
                                                 &                                                       & IMDB                       &                          &  DLRM (Acc): 87.61/87.79/85.33/88.93                                                  &                                       &                             & 6k/62k/23k/16k                     \\ \cmidrule(lr){3-3} \cmidrule(lr){5-5} \cmidrule(lr){8-8} 
                                                 &                                                       & Criteo Ad Kaggle           &                          &  BiLSTM (Acc): 78.61/78.64/78.67/78.72                                    &                                       &                             & 564k/437k/154k/200k                \\ \cmidrule(lr){2-8} 
                                                 & \multirow{3}{*}{ARD-HU\cite{hawkins2022towards}  }                     & MNIST                      & \multirow{3}{*}{-}       &  MLP (Acc): 97.98/98.30/97.04/98.23                                                   & \multirow{3}{*}{-}                    & \multirow{3}{*}{-}          & 7k/91k/4k/5k                       \\ \cmidrule(lr){3-3} \cmidrule(lr){5-5} \cmidrule(lr){8-8} 
                                                 &                                                       & IMDB                       &                          &  DLRM (Acc): 87.54/88.01/85.82/88.78                                                  &                                       &                             & 6k/58k/19k/14k                     \\ \cmidrule(lr){3-3} \cmidrule(lr){5-5} \cmidrule(lr){8-8} 
                                                 &                                                       & Criteo Ad Kaggle           &                          &  BiLSTM (Acc): 78.57/78.62/78.63/78.73                                    &                                       &                             & 571k/402k/160k/164k                \\ \cmidrule(lr){2-8} 
                                                 & \cite{hawkins2021bayesian}                            & CIFAR-10                   & -                        &  ResNet-110 (Acc): 90.4                                                         & -                                     & -                           & 7.4$\times$                              \\ \cmidrule(lr){2-8} 
                                                 & \multirow{2}{*}{\cite{kodryan2020mars}}                       & MNIST                      & \multirow{2}{*}{-}       &  LeNet-5 (Acc): 99.0                                                              & 1.19$\times$                                & \multirow{2}{*}{-}          & 10$\times$                               \\ \cmidrule(lr){3-3} \cmidrule(lr){5-6} \cmidrule(lr){8-8} 
                                                 &                                                       & CIFAR-10                   &                          &  ResNet-110 (Acc): 90.7/91.1                                              & -                                     &                             & 7$\times$/5.5$\times$                          \\ \cmidrule(lr){2-8} 
                                                 & \multirow{2}{*}{\cite{javaheripi2021autorank,samragh2019autorank}} & \multirow{2}{*}{ImageNet} & \multirow{2}{*}{ARM-A57} &  AlexNet (Top5 acc): 81.01/80.37/79.98                                                   & 543M/349M/277M                       & 4.39s/2.21s/1.61s           & \multirow{2}{*}{-}                 \\ \cmidrule(lr){5-7}
                                                 &                                                       &                            &                          &  VGG-16 (Top5 acc): 90.05/89.61/88.89                                                    & 4950M/3170M/2360M                     & 72.15s/42.55s/30.48s        &                                    \\ \cmidrule(lr){2-8}
                                                 \multirow{2}{*}{} & \multirow{2}{*}{\cite{DBLP:conf/iccv/MaTBSCB19}} & \multirow{2}{*}{ImageNet} & \multirow{2}{*}{-} & ResNet18(Top1/Top5 acc): 68.16/88.15 & \multirow{2}{*}{-} & \multirow{2}{*}{-} & 3.15M \\ \cmidrule(lr){5-5} \cmidrule(lr){8-8} 
                  &                                                                   &                           &                    & ResNet50(Top1/Top5 acc): 74.83/92.28 &                    &                    & 4.49M \\ \bottomrule
                                                 
\end{tabular}

%% file: table/KD.tex
\fontsize{6pt}{8pt}\selectfont
\caption{Summary of Automated Knowledge Distillation Studies. The ``hardware'' column indicates on which the achieved models are evaluated. The model name (e.g.,  Inception-ResNet-V2) in the ``Accuracy'' column indicates the teacher model; if the teacher is not specified, the teacher is searched during training. The latency is reported per input; otherwise is specified in the relevant table cells (e.g., s/batch). `-' indicates unavailable records.}
\begin{tabular}{@{}c|c|c|c|c|c|c|c|c@{}}
\toprule
Method                                             & Reference                                                            & Dataset                     & Hardware                                   & Accuracy                                                                        & FLOPs                  & Latency                         & \# of Parameters       & Memory                 \\ \midrule
\multirow{40}{*}{\rotatebox[origin=c]{90}{\small{Automated Knowledge Distillation}}} & \multirow{2}{*}{\cite{DBLP:conf/cvpr/LiuJTVZGW20}}  & \multirow{2}{*}{ImageNet}   & \multirow{2}{*}{Pixel 1 phone}             &  Inception-ResNet-v2 (Top5 acc): 87.5/89.1/93.1              & \multirow{2}{*}{-}     & \multirow{2}{*}{15ms/25ms/75ms} & \multirow{2}{*}{-}     & \multirow{2}{*}{-}     \\ \cmidrule(lr){5-5}
                                                   &                                                                      &                             &                                            &  Inception-ResNet-v2 (Top1 acc): 66.5/69.6/75.5              &                        &                                 &                        &                        \\ \cmidrule(lr){2-9} 
                                                   & \multirow{2}{*}{\cite{guo2021rosearch}}             & SST-2                       & \multirow{2}{*}{-}                         &  BERT (Acc): 84.1                                         & \multirow{2}{*}{-}     & \multirow{2}{*}{-}              & 24M                    & \multirow{2}{*}{-}     \\ \cmidrule(lr){3-3} \cmidrule(lr){5-5} \cmidrule(lr){8-8}
                                                   &                                                                      & Ag News                     &                                            &  BERT (Acc): 90.3                                         &                        &                                 & 16.9M                  &                        \\ \cmidrule(lr){2-9} 
                                                   & \cite{DBLP:journals/algorithms/LeiYJC21}            & Market-1501                 & -                                          &  ResNet50 (mAP/Rank): 94.7/95.6                           & -                      & -                               & 14.3M                  & -                      \\ \cmidrule(lr){2-9} 
                                                   & \multirow{3}{*}{\cite{eyono2021autokd}}             & CIFAR100                    & \multirow{3}{*}{-}                         &  Inception-Resnet-V2 (Acc): 81.2                          & \multirow{3}{*}{-}     & \multirow{3}{*}{-}              & 4M                     & \multirow{3}{*}{-}     \\ \cmidrule(lr){3-3} \cmidrule(lr){5-5} \cmidrule(lr){8-8}
                                                   &                                                                      & MIT67                       &                                            &  DARTS (Acc): 76.0                                        &                        &                                 & 6M                     &                        \\ \cmidrule(lr){3-3} \cmidrule(lr){5-5} \cmidrule(lr){8-8}
                                                   &                                                                      & ImageNet                    &                                            &  Inception-Resnet-V2 (Acc): 78.0                          &                        &                                 & 6M                     &                        \\ \cmidrule(lr){2-9} 
                                                   & \multirow{2}{*}{\cite{zhang2022autodistill}}        & GLUE                        & \multirow{2}{*}{TPUv4i}                    &  IB-$\text{BERT}_\text{LARGE}$ (AvgAcc): 80.38/81.69                  & \multirow{2}{*}{-}     & 0.45ms/0.58ms                   & 20.6M/28.5M            & \multirow{2}{*}{-}     \\ \cmidrule(lr){3-3} \cmidrule(lr){5-5} \cmidrule(lr){7-8}
                                                   &                                                                      & SQuAD                       &                                            &  IB-$\text{BERT}_\text{LARGE}$ (F1): 88.4/88.1                        &                        & 0.59ms/0.49ms                   & 22.8M/20.6M            &                        \\ \cmidrule(lr){2-9} 
                                                   & \multirow{3}{*}{\cite{chen2021scene}}               & RetailRocket                & \multirow{3}{*}{-}                         &  NextItNet (MRR@5/HR@5/NDCG@5): 0.7345/0.7964/0.7500      & \multirow{3}{*}{-}     & \multirow{3}{*}{-}              & \multirow{3}{*}{-}     & \multirow{3}{*}{-}     \\ \cmidrule(lr){3-3} \cmidrule(lr){5-5}
                                                   &                                                                      & 30Music                     &                                            &  NextItNet (MRR@5/HR@5/NDCG@5): 0.6343/0.7151/0.6544      &                        &                                 &                        &                        \\ \cmidrule(lr){3-3} \cmidrule(lr){5-5}
                                                   &                                                                      & ML-2K                       &                                            &  NextItNet (MRR@5/HR@5/NDCG@5): 0.4489/0.6519/0.4995      &                        &                                 &                        &                        \\ \cmidrule(lr){2-9} 
                                                   & \multirow{3}{*}{\cite{DBLP:conf/icpr/ZhangZYGX20}}  & \multirow{3}{*}{GLUE}       & \multirow{3}{*}{-}                         &  $\text{BERT}_\text{base}$ (SST-2/MRPC/QQP/MNLI-m/MNLI-mm/QNLI/RTE):  & \multirow{3}{*}{-}     & \multirow{3}{*}{-}              &                        & \multirow{3}{*}{-}     \\
                                                   &                                                                      &                             &                                            & 92.2/86.3/70.4/81.0/80.2/88.6/65.9                                              &                        &                                 & 42.4M                  &                        \\ \cmidrule(lr){5-5} \cmidrule(lr){8-8}
                                                   &                                                                      &                             &                                            & 86.9/79.3/67.5/76.1/75.5/83.9/58.9                                              &                        &                                 & 33.2M                  &                        \\ \cmidrule(lr){2-9} 
                                                   & \multirow{3}{*}{\cite{DBLP:conf/cvpr/LiLDLZH20}}    & \multirow{2}{*}{Cityscapes} & \multirow{2}{*}{-}                         &  Pix2Pix (mIoU): 41.71                                    & 5450M                  & \multirow{3}{*}{-}              & 0.89M                  & \multirow{3}{*}{-}     \\ \cmidrule(lr){5-6} \cmidrule(lr){8-8}
                                                   &                                                                      &                             &                                            &  GauGAN (mIoU): 61.17                                     & 31200M                  &                                 & 20.2M                  &                        \\ \cmidrule(lr){3-3} \cmidrule(lr){5-6} \cmidrule(lr){8-8}
                                                   &                                                                      & COCO-Stuff                  &                                            &  GauGAN (mIoU/FID): 35.34/25.06                           & 35400M                 &                                 & 26M                    &                        \\ \cmidrule(lr){2-9} 
                                                   & \multirow{4}{*}{\cite{cheng2021mfagan}}             & Set5                        & \multirow{4}{*}{NVIDIA V100}               &  MFANet (PSNR/LPIPS): 30.16/0.0571                        & \multirow{6}{*}{8410M} & \multirow{6}{*}{21.9ms}         & \multirow{6}{*}{0.55M} & \multirow{6}{*}{0.52G} \\ \cmidrule(lr){3-3} \cmidrule(lr){5-5}
                                                   &                                                                      & Set14                       &                                            &  MFANet (PSNR/LPIPS): 26.69/0.113                         &                        &                                 &                        &                        \\ \cmidrule(lr){3-3} \cmidrule(lr){5-5}
                                                   &                                                                      & B100                        &                                            &  MFANet (PSNR/LPIPS): 25.33/0.1332                        &                        &                                 &                        &                        \\ \cmidrule(lr){3-3} \cmidrule(lr){5-5}
                                                   &                                                                      & Urban100                    &                                            &  MFANet (PSNR/LPIPS): 24.23/0.1132                        &                        &                                 &                        &                        \\ \cmidrule(lr){2-9} 
                                                   & \multirow{2}{*}{\cite{vo2022ppcd}}                  & \multirow{2}{*}{ImageNet}   & AMD Ryzen 9 3900X                          & \multirow{2}{*}{ BigGAN (IS/FID/LPIPS): 83.13/12.76/0.62} & \multirow{2}{*}{1600M}  & 2.05s/batch                     & \multirow{2}{*}{13.6M} & \multirow{2}{*}{-}     \\ \cmidrule(lr){4-4} \cmidrule(lr){7-7}
                                                   &                                                                      &                             & NVIDIA GTX3090                             &                                                                                 &                        & 1.19s/batch                     &                        &                        \\ \cmidrule(lr){2-9} 
                                                   & \multirow{4}{*}{\cite{DBLP:conf/kdd/Xu0LS0QL21}}    & \multirow{2}{*}{GLUE}       & \multirow{4}{*}{-}                         &  $\text{BERT}_\text{base}$ (MNLI/QQP/QNLI/CoLA/SST-2/STS-B/RTE/MRPC) & \multirow{4}{*}{-}     & \multirow{4}{*}{-}              & \multirow{4}{*}{60M}   & \multirow{4}{*}{-}     \\
                                                   &                                                                      &                             &                                            & 84.1/88.8/91.2/50.5/92.6/86.9/72.7/86.4                                         &                        &                                 &                        &                        \\ \cmidrule(lr){3-3} \cmidrule(lr){5-5}
                                                   &                                                                      & SQuADv1.1                   &                                            &  $\text{BERT}_\text{base}$(EM/F1): 81.2/88.3                         &                        &                                 &                        &                        \\ \cmidrule(lr){3-3} \cmidrule(lr){5-5}
                                                   &                                                                      & SQuADv2.0                   &                                            &  $\text{BERT}_\text{base}$(EM/F1): 73.9/77.1                         &                        &                                 &                        &                        
                                                    \\ \bottomrule
\end{tabular}

%% file: table/AP.tex
\fontsize{5pt}{8pt}\selectfont
\caption{Summary of Automated Pruning Studies. The ``hardware'' column indicates on which the achieved models are evaluated. The model name (e.g., LeNet-300-100) in the ``Accuracy'' column indicates the original model before pruning; if the original model is not specified, it is searched during training. The latency is reported per input; otherwise is specified in the relevant table cells (e.g., FPS (frame per second)). $n\%$ indicates the compressed model is $n$ per cent of its original model; $\downarrow n\%$ indicates the compressed model decreases $n$ per cent compared to its original model. $n\times$ indicates that the original model is $n$ times its compressed counterpart. `-' indicates unavailable records.}

\begin{tabular}{@{}c|c|>{\centering}p{1.2cm}|>{\centering}p{1.2cm}|c|c|c|c|c@{}}
\toprule
Method                              & Reference                                                           & Dataset                    & Hardware                                    & Accuracy                                                                             & FLOPs                                                                                              & Latency                                                                                             & \# of Parameters                                                                                   & Memory                                                    \\ \midrule
\multirow{80}{*}{\rotatebox[origin=c]{90}{\small{Automated Pruning}}} & \multirow{4}{*}{\cite{zeng2021network}}            & MNIST                      & \multirow{4}{*}{-}                          & LeNet-300-100/LeNet-5(Top1 error): $\uparrow$0.08/$\uparrow$0.04                           & \multirow{2}{*}{-}                                                                                 & \multirow{4}{*}{-}                                                                                  & 77$\times$/200$\times$                                                                 & \multirow{4}{*}{-}                                        \\
\bottomrule 
\end{tabular}

%% file: table/AP_cont.tex
\fontsize{5pt}{8pt}\selectfont
\caption{Summary of Automated Pruning Studies (cont.).}
\begin{tabular}{@{}c|c|>{\centering}p{1cm}|>{\centering}p{1.1cm}|c|c|c|c|c@{}}
\toprule
Method                              & Reference                                                           & Dataset                    & Hardware                                    & Accuracy                                                                             & FLOPs                                                                                              & Latency                                                                                             & \# of Parameters                                                                                   & Memory                                                    \\ \midrule
\multirow{35}{*}{\rotatebox[origin=c]{90}{\small{Automated Pruning}}} 
& \multirow{5}{*}{\cite{guan2020dais}}               & \multirow{2}{*}{ImageNet}  & \multirow{2}{*}{Galaxy S9}                  & ResNet18/ResNet34/ResNet50(Top1 acc): 67.56/72.77/74.45                              & \multirow{2}{*}{1030M/2130M/1830M}                                                                 & \multirow{2}{*}{190ms/310ms/310ms}                                                                  & \multirow{5}{*}{-}                                                                                & \multirow{5}{*}{-}                                        \\
\bottomrule
\end{tabular}

%% file: table/AQ.tex
\fontsize{5.7pt}{8pt}\selectfont
\caption{Summary of Automated Quantization Studies. The ``hardware'' column indicates on which the achieved models are evaluated. The model name (e.g., MobileNetV1) in the ``Accuracy'' column indicates the original model before quantization. The latency is reported per input. $n\times$ indicates that the original model is $n$ times its quantized counterpart. `-' indicates unavailable records.}
\begin{tabular}{@{}c|c|>{\centering}p{1.4cm}|c|c|c|c|c@{}}
\toprule
Method                                   & Reference                                                                           & Dataset                         & Hardware                                        & Accuracy                                                                                   & Bitwdith                      & Latency                                   & Model Size                                                                       \\ \midrule
\multirow{45}{*}{\rotatebox[origin=c]{90}{\small{Automated Quantization}}} & \multirow{25}{*}{\cite{wang2019haq,DBLP:journals/ijcv/WangLLLH20}} & \multirow{25}{*}{ImageNet}      & & MobileNetV1 (Top1 acc): 67.40/70.58/71.20                                                   & \multirow{25}{*}{-}           & \multirow{2}{*}{45.51ms/57.70ms/70.35ms}  & \multirow{15}{*}{-}                                                              \\ \cmidrule(lr){5-5}
                                         &                                                                                     &                                 &  BISMO on                                                & MobileNetV1 (Top5 acc): 87.90/89.77/90.19                                                   &                               &                                           &                                                                                  \\ \cmidrule(lr){5-5} \cmidrule(lr){7-7}
                                         &                                                                                     &                                 &   Xilinx Zynq-7020                                              & MobileNetV2 (Top1 acc): 66.99/70.90/71.89                                                   &                               & \multirow{2}{*}{52.12ms/66.92ms/82.34ms}  &                                                                                  \\ \cmidrule(lr){5-5}
                                         &                                                                                     &                                 &                                                 & MobileNetV2 (Top5 acc): 87.33/89.91/80.36                                                   &                               &                                           &                                                                                  \\ \cmidrule(lr){4-5} \cmidrule(lr){7-7}
                                         &                                                                                     &                                 &             & MobileNetV1 (Top1 acc): 65.33/69.97/71.20                                                   &                               & \multirow{2}{*}{57.40ms/77.49ms/99.66ms}  &                                                                                  \\ \cmidrule(lr){5-5}
                                         &                                                                                     &                                 &    BISMO on                                             & MobileNetV1 (Top5 acc): 86.60/89.37/90.08                                                   &                               &                                           &                                                                                  \\ \cmidrule(lr){5-5} \cmidrule(lr){7-7}
                                         &                                                                                     &                                 & Xilinx VU9P                                                & MobileNetV2 (Top1 acc): 67.01/69.45/71.85                                                   &                               & \multirow{2}{*}{73.97ms/99.07ms/127.03ms} &                                                                                  \\ \cmidrule(lr){5-5}
                                         &                                                                                     &                                 &                                                 & MobileNetV2 (Top5 acc): 87.46/88.94/90.24                                                   &                               &                                           &                                                                                  \\ \cmidrule(lr){4-5} \cmidrule(lr){7-7}
                                         &                                                                                     &                                 & \multirow{2}{*}{BitFusion}                      & MobileNetV1 (Top1 acc): 67.45/70.40/70.90                                                   &                               & \multirow{2}{*}{7.86ms/11.09ms/19.98ms}   &                                                                                  \\ \cmidrule(lr){5-5}
                                         &                                                                                     &                                 &                                                 & MobileNetV1 (Top5 acc): 87.85/89.69/89.95                                                   &                               &                                           &                                                                                  \\ \cmidrule(lr){4-5} \cmidrule(lr){7-8} 
                                         &                                                                                     &                                 & \multirow{8}{*}{-}                              & MobileNetV1 (Top1 acc): 57.14/67.66/71.74                                                   &                               & \multirow{8}{*}{-}                        & \multirow{2}{*}{1.09MB/1.58MB/2.07MB}                                            \\ \cmidrule(lr){5-5}
                                         &                                                                                     &                                 &                                                 & MobileNetV1 (Top5 acc): 81.87/88.21/90.36                                                   &                               &                                           &                                                                                  \\ \cmidrule(lr){5-5} \cmidrule(lr){8-8} 
                                         &                                                                                     &                                 &                                                 & MobileNetV2 (Top1 acc): 66.75/70.90/71.47                                                   &                               &                                           & \multirow{2}{*}{0.95MB/1.38MB/1.79MB}                                            \\ \cmidrule(lr){5-5}
                                         &                                                                                     &                                 &                                                 & MobileNetV2 (Top5 acc): 87.32/89.76/90.23                                                   &                               &                                           &                                                                                  \\ \cmidrule(lr){5-5} \cmidrule(lr){8-8} 
                                         &                                                                                     &                                 &                                                 & ResNet50 (Top1 acc): 70.63/75.30/76.14                                                      &                               &                                           & \multirow{2}{*}{6.30MB/9.22MB/12.14MB}                                           \\ \cmidrule(lr){5-5}
                                         &                                                                                     &                                 &                                                 & ResNet50 (Top5 acc): 89.93/92.45/92.89                                                      &                               &                                           &                                                                                  \\ \cmidrule(lr){2-8} 
                                         & \multirow{2}{*}{\cite{DBLP:conf/uai/LaceyTA18}}                    & MNIST                           & \multirow{2}{*}{-}                              & LeNet5 (Val. error): 1.14/1.69/2.32                                                         & \multirow{2}{*}{-}            & \multirow{2}{*}{-}                        & \multirow{2}{*}{-}                                                               \\ \cmidrule(lr){3-3} \cmidrule(lr){5-5}
                                         &                                                                                     & ImageNet                        &                                                 & AlexNet (Val. error): 47.46/47.69/48.54                                                     &                               &                                           &                                                                                  \\ \cmidrule(lr){2-8} 
                                         & \multirow{3}{*}{\cite{DBLP:conf/eccv/YuHLSCF20}}                   & CIFAR-10                        & \multirow{3}{*}{-}                              & ResNet20 (Top1 acc): 92.30/92.12/92.04                                                      & Avg: 3.5/3.3/2.9              & \multirow{3}{*}{-}                        & 10.19$\times$/10.74$\times$/12.08$\times$ \\ \cmidrule(lr){3-3} \cmidrule(lr){5-6} \cmidrule(lr){8-8} 
                                         &                                                                                     & \multirow{2}{*}{ImageNet}       &                                                 & ResNet50 (Top1 acc): 76.67/75.71                                                            & \multirow{2}{*}{Avg: 3.8/2.9} &                                           & \multirow{2}{*}{-}                                                               \\ \cmidrule(lr){5-5}
                                         &                                                                                     &                                 &                                                 & ResNet50 (Top5 acc): 93.55/92.83                                                            &                               &                                           &                                                                                  \\ \cmidrule(lr){2-8} 
                                         & \multirow{2}{*}{\cite{DBLP:conf/icassp/XuHYLM21}}                  & Penn Treebank Corpus            & \multirow{6}{*}{-}                              & Transformer (Perplexity): 56.82/58.23                                                       & Avg: 2.0/2.2                  & \multirow{6}{*}{-}                                       & 6.5MB/4.8MB                                                                      \\ \cmidrule(lr){3-3} \cmidrule(lr){5-6} \cmidrule(lr){8-8} 
                                         &                                                                                     & Conversational Telephone Speech &                                                 & Transformer (Perplexity): 42.39/42.75                                                       & Avg: 1.9/2.5                  &                                           & 8.0MB/9.1MB                                                                      \\ \cmidrule(lr){2-8} 
                                         & \multirow{7}{*}{\cite{sun2021effective}}                           & CIFAR-10                        & \multirow{7}{*}{-}                              & ResNet20 (Top1 acc): 92.54                                                                  & Avg: 3.04                     & \multirow{7}{*}{-}                        & \multirow{7}{*}{-}                                                               \\ \cmidrule(lr){3-3} \cmidrule(lr){5-6}
                                         &                                                                                     & \multirow{2}{*}{ImageNet}       &                                                 & ResNet18/ResNet34/ResNet50/MobileNetV2 (Top1 acc):                                          & Avg:                          &                                           &                                                                                  \\
                                         &                                                                                     &                                 &                                                 & 70.70/74.30/76.22/69.10                                                                    & 3.95/4.01/3.98/4.02           &                                           &                                                                                  \\ \cmidrule(lr){3-3} \cmidrule(lr){5-6}
                                         &                                                                                     & \multirow{4}{*}{COCO}           &                                                 & Faster R-CNN (AP/AP.5/AP.75/$\text{AP}_\text{S}$/$\text{AP}_\text{M}$/$\text{AP}_\text{L}$) & \multirow{2}{*}{Avg: 4.00}       &                                           &                                                                                  \\
                                         &                                                                                     &                                 &                                                 & 37.4/58.1/40.6/22.1/40.4/47.9                                                              &                               &                                           &                                                                                  \\ \cmidrule(lr){5-6}
                                         &                                                                                     &                                 &                                                 & RetinaNet (AP/AP.5/AP.75/$\text{AP}_\text{S}$/$\text{AP}_\text{M}$/$\text{AP}_\text{L}$)    & \multirow{2}{*}{Avg: 4.00}       &                                           &                                                                                  \\
                                         &                                                                                     &                                 &                                                 & 36.4/55.8/38.6/20.8/39.9/47.6                                                              &                               &                                           &                                                                                  \\ \cmidrule(lr){2-8} 
                                         & \multirow{2}{*}{\cite{DBLP:journals/lgrs/WeiCLX21}}                & \multirow{2}{*}{NWPU-RESISC45}  & \multirow{2}{*}{-}                              & ResNet34 (acc): 92.66/92.75/92.73                                                           & \multirow{2}{*}{-}            & \multirow{2}{*}{-}                        & 13.17MB/16.50MB/18.86MB                                                          \\ \cmidrule(lr){5-5} \cmidrule(lr){8-8} 
                                         &                                                                                     &                                 &                                                 & SqueezeNet (acc): 88.45/88.59/88.61                                                         &                               &                                           & 0.45MB/0.55MB/0.69MB                                                             \\ \cmidrule(lr){2-8} 
                                         & \multirow{2}{*}{\cite{wu2018mixed}}                                & CIFAR-10                        & \multirow{2}{*}{-}                              & ResNet20/ResNet56/ResNet110 (acc): 92.00/94.12/94.39                                        & \multirow{2}{*}{-}            & \multirow{2}{*}{-}                        & 16.6$\times$/18.93$\times$/20.3$\times$   \\ \cmidrule(lr){3-3} \cmidrule(lr){5-5} \cmidrule(lr){8-8} 
                                         &                                                                                     & ImageNet                        &                                                 & ResNet18/ResNet34 (acc): 69.58/73.37                                                        &                               &                                           & 21.1$\times$/19.0$\times$                              \\ \bottomrule
\end{tabular}

%% file: content/5.joint.tex
\section{Joint Automated Design Strategies}\label{section:joint}
Neural architecture design and compression are divergent routines to achieve efficient deep learning models and a combination of these two strategies is an intuitive way to achieve even more efficiency than merely relying on either technique individually. The most natural way in this regard is a sequential pipeline: design then compress. Under the context of design automation, the pipeline is commonly recognized as a search-compress scheme. 
There are bare studies on combining tensor decomposition and NAS. We will thus concentrate on the other three compression techniques in this section. 
TABLE \ref{tab:joint} summarizes main references on joint automated design work. 

\subsection{Joint Search and Knowledge Distillation}
For knowledge distillation, the intuitive joint automated design strategy is to first search for a powerful teacher model and then achieve a compressed student model by automated knowledge distillation. An improved strategy is FasterSeg \cite{fasterseg2020chen}, which searches for student and teacher alternatively so as to ensure optimal teacher-student matching pairs and effective knowledge distillation. Instead of searching for the entire teacher model, Guan et al. \cite{DBLP:conf/eccv/GuanZWZYBT20} target the feature distillation and search for the optimal teacher's feature aggregation for distillation. For each layer group in the teacher model, the authors try to search for the aggregation weights of each layer's feature in that group; then the features are weighted sum to guide the knowledge distillation of the corresponding student layer. This work mimics the multi-teacher distillation scheme but with a single-teacher, multi-feature distillation proposal and automated teacher knowledge aggregation.

\subsection{Joint Search and Pruning}
Different from the search-compress scheme, pruning can be incorporated into neural architecture search as integrity. The aforementioned automated pruning approaches (in Section \ref{ssec: AP}) decide the pruning strategy based on the weights of trained networks. However, a recent study \cite{DBLP:conf/iclr/LiuSZHD19} indicates that the essence of network pruning is the pruned structure instead of the weights inherited from the base network, which means that the pruning strategy can be determined before the model is trained. In light of this critical conclusion, the research direction turns to finding optimal pruned structures, e.g., the channel number in each layer, rather than to determining important channels from the base network. This objective is basically the same as NAS so numerous works emerge to use NAS to achieve pruning, where the pruning strategy and network structure are sampled together before training in the search process  \cite{DBLP:conf/iccv/LiuMZG0C019,liu2021latency,DBLP:conf/ijcai/LinJZZW020,chang2021acp,liu2021superpruner,lin2021aacp}.
However, different from pure NAS, the NAS-based pruning 1) requires an existing state-of-the-art model as the pruning base, 2) only searches the network hyperparameters (e.g., channel number, layer number) instead of the type of operators, and 3) is bounded by searching smaller hyperparameters than the base network. MetaPruning \cite{DBLP:conf/iccv/LiuMZG0C019} is an early work in this direction, which attempts to find the optimal number of channels for each layer. It first constructs a PruningNet to generate weights for pruned networks so that fine-tuning is not required during search time; then it utilizes an evolutionary algorithm to search for the optimal combination of layers with different numbers of channels. The search procedure is highly efficient and no fine-tuning is obliged after searching. However, the search space is large especially when there are many channels in the base network. Some following works thus attempt to shrink the search space by only searching channel number ratios \cite{DBLP:conf/ijcai/LinJZZW020} or using a clustering algorithm to cluster channels in each layer as the initial number of channels in the pruned network for later search \cite{chang2021acp}. Other works \cite{liu2021superpruner,lin2021aacp}, on the other hand, seek to develop a performance estimator to accelerate the searching process. 
Overall, most NAS-based pruning process is analogous to \textit{searching optimal pruning rates} but without pruning low magnitude channels at each iteration, instead, it trains from scratch with the searched structure or quickly derives the performance with an additional performance estimator.

Alternative to unifying NAS and pruning by sampling the pruning strategy and model architecture before training, SpArSe \cite{sparse2019igor} selects the pruning strategy based on a trained model but optimizes the pruning hyperparameters and model architecture jointly. Specifically, it first samples a configuration of the model architecture (e.g., operators and connectivity) and hyperparameters of pruning algorithms (i.e., Sparse Variational Dropout (SpVD) and Bayesian Compression (BC)); it then trains the model and prunes it. After evaluation, it optimizes the sampling configuration with multi-objective Bayesian optimization. In this way, both structured and unstructured pruning strategies can be optimized together with the model architecture.

\subsection{Joint Search and Quantization}
\subsubsection{Optimizing Search and quantization separately} Some researchers consider optimizing the two components separately and particularly focus on automating the architecture design process.  
For example, Cai et al. use the ProxylessNAS \cite{proxyless} framework to firstly search for an efficient architecture and then perform quantization-aware fine-tuning to further compress the model \cite{DBLP:conf/iccvw/CaiWWWLH19}; Liu et al. \cite{DBLP:conf/iccvw/LiuHSLCW21} include the quantization-friendly activation function ReLU6 in the search space to facilitate the following quantization; Peter et al. \cite{DBLP:conf/icpr/PeterRP20} compare the performance of post-training quantization (PTQ) and quantization-aware training (QAT) and indicate that the PTQ performance drops rapidly when the bitwidth is smaller than 4 while QAT only has marginal drop even with 1-bit precision. This is due to the incompatibility between quantization and the network structure, which is primarily optimized for full precision. 

In light of this, attentions are drawn to quantization-aware NAS, where the search space is based on quantized operators \cite{DBLP:conf/eccv/BulatMT20,DBLP:conf/cvpr/PhanLHSCS20,DBLP:conf/eccv/KimSC20,DBLP:conf/iccvw/ShenHXW19,kim2020frostnet}. Most of these works are for binary neural networks (BNNs) because of their extreme computation and memory savings \cite{DBLP:conf/eccv/BulatMT20,DBLP:conf/cvpr/PhanLHSCS20,DBLP:conf/eccv/KimSC20,DBLP:conf/iccvw/ShenHXW19,DBLP:conf/cvpr/XuZL0HD21}. Shen et al. \cite{DBLP:conf/iccvw/ShenHXW19} apply an evolutionary search to optimize the layer-wise channel expansion ratio for binarized VGG and ResNet. Phan et al. \cite{DBLP:conf/cvpr/PhanLHSCS20} discover that binary DWConv has limited representation capability and thus attempt to search for the group number for each layer of the MobileNet via an evolutionary algorithm as well. BATS \cite{DBLP:conf/eccv/BulatMT20} and BNAS \cite{DBLP:conf/eccv/KimSC20} concurrently emerge in the sense of differentiable quantization-aware NAS. They both design a BNNs-oriented search space discarding the separable convolution due to its failure during quantization; BNAS keeps convolutions and dilated convolutions, and BATS keeps group convolutions and dilated group convolutions.
It is interesting to note that the BATS binarizes only activations during training but keeps weights at full precision and binarizes the weights after training with a marginal drop in accuracy ($\sim$1\%). Instead of optimizing the architecture as a whole, Xu et al. \cite{DBLP:conf/cvpr/XuZL0HD21} probe layer-wise searching under the KD guidance for BNN object detectors. 

The above work only automatically optimizes the neural architectures while using fixed quantization policies. As a result, the final model has a large possibility to be sub-optimal: e.g., the BNN with the searched architecture is not necessarily smaller and more accurate than a mixed-precision model.  

\subsubsection{Optimizing search and quantization jointly} 
Jointly optimizing the neural architecture and quantization policy can find the best combination of both worlds. 
The most intuitive joint approach is to construct a new search space that contains both structure and bitwidth choices, and the search process is the same as for traditional hardware-aware NAS \cite{chen2018joint,DBLP:conf/iccad/GongJWLLP19}.
For example, JASQ \cite{chen2018joint} employs a classical evolutionary search algorithm to explore both operators and cell-wise quantization bitwidths; Gong et al. \cite{DBLP:conf/iccad/GongJWLLP19}, instead, engage differentiable neural architecture search to explore a mixed-precision supernet of MBConv hyperparameters and bitwidths. Though simple and effective, this strategy is inefficient: the search space is enlarged by many times that JASQ requires three GPU days to search for a suitable model for each given resource budget.

To tackle this issue, some studies decouple the supernet training and search process so that no training is required in the search process \cite{DBLP:conf/eccv/GuoZMHLWS20,DBLP:conf/cvpr/WangWCLL0LH20,DBLP:conf/iccv/ShenLGLL0YYO21}. The weight-sharing training paradigm is further adopted to reduce the training cost. 
SPOS \cite{DBLP:conf/eccv/GuoZMHLWS20} designs a single path weight-sharing strategy that trains a supernet independently to the search process.
In each iteration of the supernet training, only one single path of the supernet is activated and trained. The single-path activation is realized by uniformly sampling one candidate block at each block level of the supernet. The weights of the same block are shared among different supernet training iterations (i.e., different paths). In the search process, an evolutionary algorithm is performed on randomly sampled paths of the pre-trained supernet to find the optimal combination of architecture and block-wise bitwidths. As the supernet is well pre-trained, each architecture only performs inference during the evolutionary search. This SPOS strategy improves the efficiency of both the supernet training and search. APQ \cite{DBLP:conf/cvpr/WangWCLL0LH20} adopts the once-for-all \cite{cai2020once} supernet training strategy, which requires little time for retraining the searched model but is inefficient to support different bitwidths in the supernet. Therefore, the authors devise a quantization-aware accuracy predictor so that they can train the once-for-all supernet in full precision and derive the accuracy of a quantized network via the accuracy predictor. However, it is hard and time-consuming to collect a large volume of [quantized network, accuracy] pairs for preparing the quantization-aware accuracy predictor due to lengthy quantization-aware training. To alleviate this issue, the authors first train a full precision accuracy predictor with abundant [full-precision network, accuracy] pairs from the supernet, and then fine-tune it with only a few [quantized network, accuracy] pairs to obtain its quantization-aware counterpart. Next, they then do a simple evolutionary search with the accuracy predictor to achieve a new model for a given resource budget. Instead of using the accuracy predictor to deal with the different bitwidths problem, Shen et al. \cite{DBLP:conf/iccv/ShenLGLL0YYO21} conceive the once quantization-aware (OQA) training strategy, where a set of quantized supernets with sequential bitwidths are created via the bit inheritance scheme: the $k-1$ bit supernet uses double quantization step size of its $k$ bit counterpart. The authors experimentally show that with just one epoch fine-tuning the lower-bit network with bit inheritance outperforms its counterpart with QAT from scratch. The final architecture is derived by a simple coarse-to-fine architecture selection procedure without any retraining. However, OQA only supports fixed precision quantization policies as well. 

The most recent work, QFA \cite{bai2021batchquant}, manages to allow joint mixed-precision quantization and architecture search without retraining. The authors propose a new quantization strategy, batch quantization, to stabilize the estimation of the \textit{scale factor} so the mixed-precision supernet can be stably trained. They replace the progressive shrinking training strategy in the work \cite{DBLP:conf/iccv/ShenLGLL0YYO21} with a newly proposed two-stage training strategy reducing more than half training epochs.

\begin{table*}[]
    \centering
    \input{table/joint}
    \label{tab:joint}
\end{table*}

\begin{table*}[]
\ContinuedFloat 
    \centering
    \input{table/joint_cont}
    \label{tab:jointcont}
\end{table*}

%% file: table/joint.tex
\fontsize{5pt}{8pt}\selectfont
\caption{Summary of Joint Automated Design Strategies. The ``hardware'' column indicates on which the achieved models are evaluated. The model name (e.g., MobileNetV1) in the ``Accuracy'' column indicates the base model used for search; if the original model is not specified, it is searched during training. The latency is reported per input; otherwise is specified in the relevant table cells (e.g., FPS). $\downarrow n\%$ indicates the compressed model decreases $n$ per cent compared to its original model. $\sim$ indicates the exact data is not reported in the paper and thus estimated from the reported figures. `-' indicates unavailable records.}
\begin{tabular}{@{}c|c|>{\centering}p{1cm}|>{\centering}p{1.1cm}|c|c|c|c|c@{}}
\toprule
Method                              & Reference                                                           & Dataset                    & Hardware                                    & Accuracy                                                                             & FLOPs                                                                                              & Latency                                                                                             & \# of Parameters                                                                                   & Memory                                                    \\ \midrule

\multirow{8}{*}{\rotatebox[origin=c]{90}{\scriptsize{Joint Search and KD}}} & \multirow{2}{*}{\cite{DBLP:conf/eccv/GuanZWZYBT20}} & CIFAR100                    & \multirow{2}{*}{-}                         & (Acc): 79.74/78.18/75.85                                                        & \multirow{2}{*}{-}     & \multirow{2}{*}{-}              & 2.77M/1.47M/0.7M       & \multirow{2}{*}{-}     \\ \cmidrule(lr){3-3} \cmidrule(lr){5-5} \cmidrule(lr){8-8}
                                                   &                                                                      & CINIC-10                    &                                            & (Acc): 85.41/79.45/79.38                                                        &                        &                                 & 1.27M/0.36M/0.36M      &                        \\ \cmidrule(l){2-9} 
                                                   & \multirow{3}{*}{\cite{fasterseg2020chen}}           & Cityscapes                  & & (mIoU): 71.5                                                                    & \multirow{3}{*}{-}     & 163.9 FPS                       & \multirow{3}{*}{-}     & \multirow{3}{*}{-}     \\
\bottomrule
\end{tabular}

%% file: table/joint_cont.tex
\fontsize{5pt}{8pt}\selectfont
\caption{Summary of Joint Automated Design Strategies (cont.).}
\begin{tabular}{@{}c|c|c|c|c|c|c|c|c@{}}

\toprule
Method                    & Reference                                                            & Dataset                   & Hardware                                      & Accuracy                                        & Bitwdith               & Latency            & FLOPs                        & Model Size              \\ \midrule

\multirow{25}{*}{\rotatebox[origin=c]{90}{\scriptsize{Joint Search and Quantization}}}   & \multirow{2}{*}{\cite{DBLP:conf/eccv/KimSC20}}      & ImageNet                  & \multirow{2}{*}{-}                            & Top1/Top5 acc: 63.51/83.91                      & \multirow{2}{*}{1}     & \multirow{2}{*}{-} & 90M                          & \multirow{2}{*}{-}      \\ \cmidrule(lr){3-3} \cmidrule(lr){5-5} \cmidrule(lr){8-8}
                                                &                                                                      & CIFAR10                   &                                               & Top1 acc: 94.43                                 &                        &                    & 656M                         &                         \\ \cmidrule(lr){2-9} 
                                                & \multirow{2}{*}{\cite{DBLP:conf/iccvw/ShenHXW19}}   & ImageNet                  & \multirow{2}{*}{-}                            & ResNet18(Top1/Top5 acc): 69.65/89.08            & \multirow{2}{*}{1}     & \multirow{2}{*}{-} & 660M                         & \multirow{2}{*}{-}      \\ \cmidrule(lr){3-3} \cmidrule(lr){5-5} \cmidrule(lr){8-8}
                                                &                                                                      & CIFAR10                   &                                               & VGG-small(Top1 acc): 93.06                      &                        &                    & 59.3M                        &                         \\
\bottomrule
\end{tabular}

%% file: content/6.future.tex
\section{Future Directions}\label{sec:future}
The design automation for efficient deep learning models is an exciting area yet still in its infant research phase. Existing work has covered almost its every research aspect and laid firm foundations for its future evolution. In this section, we discuss several future directions that are worthy to delve into. 

\subsection{Beyond Computer Vision}
Most current works validate their performance on computer vision (CV) tasks, e.g., image classification and object detection. This is ascribable to two reasons: 1) the CV area is the most developed area regarding deep learning research that abundant resources are available and experimental protocol is well standardized; thus it requires only bare efforts to prepare comparison benchmarks; 2) the CV models are mainly based on the CNN family and much manual engineering has been devoted to devising efficient CNN models so it is relatively easy to establish an efficient search space or base models for compression. Nevertheless, CV is not the only field that expects efficient deep learning models; applications like IoT are more resource-constrained but seldom investigated regarding the design automation for efficient deep learning. Since these domains have dissimilar characteristics to CV, specific research is desired to go beyond the CV. Though some works have already been presented, like \cite{guo2021rosearch,zhang2022autodistill,DBLP:conf/icassp/XuHYLM21} on automatically compressing language models, these are very limited and more room for different applications and the deeper investigation remains to be explored. 

\subsection{Accelerating Search}
While searched models are driven to be efficient, the search process is often time- and computation-intensive because vast numbers of candidate models need to be evaluated before choosing the best one. This is problematic as the search process is task- and/or hardware-specific and thus a new search is required when a new task or hardware arises. Accelerating the search process can not only facilitate the task/hardware adaptation but also save the energy consumed during the search. The weight-sharing paradigm \cite{DBLP:conf/icml/PhamGZLD18,liu2018darts} can alleviate this problem but this is still too slow, especially for the hardware-aware settings. Using proxies is another strategy to accelerate the search process, which approximates different models' performance without fully training them \cite{DBLP:conf/aaai/RealAHL19}. Most recently, \textit{zero-cost} proxies for NAS have attained great interest and shown outstanding performance on some NAS benchmarks \cite{DBLP:conf/iclr/AbdelfattahMDL21}. This strategy aims to rank network architectures without any training or even without seeing data \cite{DBLP:conf/iccv/LinWSCS00021,DBLP:conf/iclr/AbdelfattahMDL21}. Though remarkable efforts have been made in accelerating conventional NAS, how these approaches can be efficiently and effectively applied to the hardware-aware NAS and automated compression is still an open question. For example, which proxies can be used for automated quantization? Or can we use the same proxies for different quantization bitwidths? 

Another direction for accelerating the search is to reduce the search space. For example, EfficientMet \cite{efficientnet} fixes a baseline model and only searches the constant scale ratios of width/depth/resolution. Only a small grid search can produce extraordinary performance. An effective and reduced search space is based on extensive investigation and human knowledge and thus demands considerable effort in this direction. 

\subsection{Hardware Simulation}
The feedback (e.g., latency and memory usage) from target devices is a necessity for exploring efficient deep learning models automatically. However, running an enormous amount of models on target devices is labour- and time-intensive, especially when the devices are general-purpose computing systems (e.g., MCU) instead of specific NN accelerators (e.g., TPUs). Furthermore, the frequent communication between the target devices and the devices, on which the design program runs, takes even more time. Therefore, it is desirable to have an accurate hardware simulator that simulates the behaviour of DNN hardware implementations. This simulator would accelerate the automated design process by running all models and obtaining all feedback on a powerful server. In addition, with a hardware simulator, beginners do not need to have cross-disciplinary knowledge in the hardware setting and compilation to collect hardware-cost data; the design process becomes easy to get started with. Although some works have reported hardware simulators like \cite{DBLP:conf/cvpr/GholamiKWTYJZK18,DBLP:conf/glvlsi/SunZWCL19}, the precision and hardware information that can be provided are far from satisfactory. It is also attractive to have a general software platform that offers the simulation of deploying DNNs on diverse and configurable devices. 

\subsection{Benchmarking}
Despite tremendous results that have been reported to achieve state-of-the-art performance, most of these works are not based on the same protocol thus resulting in skewed comparison. First, some literature uses direct evaluation metrics while others use indirect ones. Superiority on indirect metrics does not necessarily mean more efficiency on real devices. Second, the comparison results can vary on different hardware or different implementation details on the same hardware. Finally, the dataset/task settings also influence the performance significantly. For example, even using the same dataset, different input resolutions would lead to different hardware consumption. It is thus imperative to benchmark these evaluation settings to resolve unfair comparisons and make this domain more reproducible.  In addition, a unified benchmark dataset can also facilitate the comparison of design automation algorithms. Li et al. \cite{DBLP:conf/iclr/LiYFZZYY0HL21} recently develop a comprehensive benchmark dataset of hardware-aware NAS on three categories of hardware (e.g., commercial edge devices, FPGA, and ASIC). However, this benchmark dataset primarily aims to facilitate the development of search strategies, and therefore is relatively small and based on only two spaces. There so far lacks well-recognized benchmark settings and a thorough dataset in this domain. 

Apart from the above less noticed and underdeveloped directions, other directions like developing novel efficient operators and competent design algorithms have been widely researched, but there is still large room to improve.

%% file: content/7.conclusion.tex
\section{Conclusion}\label{sec:conclusion}
This survey provides a thorough review of the design automation techniques for fast, lightweight and effective deep learning models.
We analyze and summarize current studies into three categories: 1) by searching, 2) by compressing, and 3) by jointly. In addition, we conclude the evaluation metrics specific to efficient deep learning models.
At the end of this work, we afford discussion of existing issues and future directions for both novices and experienced researchers.